\documentclass[multiauthors,onecolumn]{LaTeXclass/cohere}

\usepackage{latexsym}
\usepackage{booktabs}
\usepackage{ragged2e}
\usepackage{lipsum}
\usepackage{pdfpages}
\usepackage{colortbl}
\usepackage{tabulary}
\usepackage{longtable}
\usepackage{array}
\usepackage{placeins}
\usepackage{tablefootnote}
\usepackage{tcolorbox}
\usepackage{wrapfig}
\usepackage{breqn}

\usepackage{pifont}
\definecolor{deeppurple}{HTML}{9e02f7}
\definecolor{forestgreen}{HTML}{2e7d43}
\usepackage{multirow}
\usepackage{tabularx}

\usepackage{enumitem}
\usepackage{arydshln}
\usepackage{amsmath}
\usepackage{amsfonts}
\usepackage{cleveref}
\usepackage{wrapfig}
\usepackage{tcolorbox}
\usepackage{mdframed}

\usepackage{fancyvrb}
\usepackage{fvextra}
\usepackage{subcaption}
\usepackage{microtype}
\usepackage{bbm}

\usepackage[utf8]{inputenc}

\usepackage{arydshln}

\usepackage{xspace} 
\usepackage{makecell} 
\usepackage{longtable}
\usepackage{array}
\usepackage{siunitx}     
\usepackage{nicematrix}
\usepackage{comment}
\usepackage{amssymb}
\usepackage{listings}

\definecolor{english}{rgb}{0.12156862745098039    , 0.4666666666666667    , 0.7058823529411765    }
\definecolor{nonenglish}{rgb}{1.0 , 0.4980392156862745 , 0.054901960784313725  }

\usepackage{newunicodechar}
\usepackage{tcolorbox}

\usepackage{twemojis}

\usepackage{caption}
\lstdefinestyle{mystyle}{
  basicstyle=\ttfamily\small,
  breaklines=true,
  frame=single,
  backgroundcolor=\color{gray!5},
  captionpos=b,
  aboveskip=1em,
  belowskip=1em
}
\newcommand{\english}{\textcolor{english}{$\blacksquare$}}
\newcommand{\nonenglish}{\textcolor{nonenglish}{$\blacksquare$}}

\setcitestyle{number,square}

\title{When Life Gives You Samples}
\subtitle{The Benefits of Scaling up Inference Compute for Multilingual LLMs}

\author{name={Ammar Khairi\fa},affiliation={1}}
\author{name={Daniel D'souza},affiliation={1}}
\author{name={Ye Shen},affiliation={2}}
\author{name={Julia Kreutzer\psa},affiliation={1}}
\author{name={Sara Hooker\psa},affiliation={1}}
\affiliations{
    \item[1] Cohere Labs
    \item[2] Cohere
}

\corresponding[*]{\{\url{ammar}, \url{juliakreutzer}, \url{sarahooker}\}\url{{@cohere.com}}}

\abstract{
\justifying
Recent advancements in large language models (LLMs) have shifted focus toward scaling inference-time compute—improving performance without retraining the model. A common approach is to sample multiple outputs in parallel, and select one of these as the final output. However, work to date has focused on English and a handful of domains such as math and code. In contrast, we are most interested in techniques that generalize across open-ended tasks, formally verifiable tasks, and across languages. In this work, we study how to robustly scale inference-time compute for open-ended generative tasks in a multilingual, multi-task setting. 

Our findings show that both sampling strategy---based on temperature variation---and selection strategy must be adapted to account for diverse domains and varied language settings. We evaluate existing selection methods, revealing that strategies effective in English often fail to generalize across languages. We propose novel sampling and selection strategies specifically adapted for multilingual and multi-task inference scenarios, and show they yield notable gains across languages and tasks. In particular, our combined sampling and selection methods lead to an average +6.8 jump in win-rates for our 8B models on m-ArenaHard-v2.0 prompts, against proprietary models such as Gemini. At larger scale, Command-A (111B model) equipped with our methods, shows +9.0 improvement in win-rates on the same benchmark with just five samples against single-sample decoding, a substantial increase at minimal cost.  Our results underscore the need for language- and task-aware approaches to inference-time compute, aiming to democratize performance improvements in underrepresented languages.
}

\renewcommand{\FONTauthors}{\bfseries\Large}
\renewcommand{\FONTauthor}{\bfseries\Large}
\begin{document}

\section{Introduction}\label{sec:introduction}
Traditionally, if you wanted higher performance from a machine learning model, you paid for it with more training or data or parameters. A key departure from this is the recent emphasis on scaling up compute at inference time rather than at training time~\citep{wu2024scaling,hooker2024limitationscomputethresholdsgovernance,snell2025scaling}. The combination of growing generative capabilities of large language models (LLMs) paired with better sampling techniques has spurred progress in inference-time compute strategies. These strategies allow for improvements in performance by spending more compute without any alterations to the model itself. 
However, much remains unknown about how to best search for optimal solutions using inference compute alone, especially for open-ended generative tasks~\citep{zhang2025and}. Even less established is how to tailor inference compute strategies to languages beyond English,  which are traditionally under-served by state-of-the-art systems~\citep{ustun2024aya,dang2024aya,dash2025ayavisionadvancingfrontier}, and underrepresented in LLM research.

In our work, our goal is to understand how to \emph{most robustly scale inference compute across languages for generative tasks}. For a given model, how can we best invest a fixed budget of inference-time compute to improve performance across all languages? 
We are most interested in techniques that generalize across open-ended tasks and formally verifiable tasks, and across languages. Hence, our setting is \emph{extremely multi-task} with many different performance constraints to balance.

We focus on \emph{parallel scaling}~\citep{wang2023selfconsistency, welleck2024from, zhang2025and} which increases inference-time compute by first generating multiple generations in parallel and then selecting one of the sampled outputs as final output.\footnote{This idea is known under many other names, e.g. Best-of-$N$~\citep{huang2025best}, ``repeated-sampling-then-voting''~\citep{chen2025we}, or rejection sampling~\citep{touvron2023llama}.} You can think of \emph{parallel scaling} as an endeavor to make the best lemonade from an already grown lemon tree (the trained model). This requires two steps, \textbf{1) sampling:} strategically choosing which lemons and how many lemons to harvest, \textbf{2) selection:} carefully evaluating each to isolate a lemon within that harvest that produces the best lemonade. These two stages, and and how well they are aligned with the evaluation metric of interest, determine the power of inference scaling ~\citep{stroebl2024inference,brown2024large,huang2025best}. In this work, we revisit both stages from first principles and exhaustively evaluate existing techniques to see which work in a massively diverse task environment. In contrast to previous work, our focus is on generalizable solutions across a diverse set of tasks (both open-ended and more studied tasks like math and machine translation), models and languages, starting from competitive baselines. Our results suggest that existing sampling and selecting strategies are not well fit to settings beyond English.

We propose novel techniques that depart from go-to solutions for English to account for a diverse language and task setting. We are interested in competitive production settings so throughout our work, we measure gains on state-of-art multilingual models which are already extremely well performing and well polished, such as Aya Expanse and Qwen3 at the 8B scale, and Command-A with 111B parameters. We rigorously compare against standard Best-of-N scoring with the strongest reward model according to prior benchmarking, and ambitiously compare performance of the 8B models against Gemini 2.0, a proprietary model. Therefore, wins in this competitive setup are hard to obtain. Our methods, thanks to being carefully crafted for the multilingual and multi-task demands with small factors of inference compute growth, achieve impressive gains across the bench, which means that the additional compute is extremely well spent. 

Our primary contributions are the following:
\begin{enumerate}[leftmargin=10pt]
    \item \textbf{Extensive study of existing methods.} 
    While many prior approaches exist for this problem~\citep{ippolito-etal-2019-comparison,shi-etal-2024-thorough,snell2025scaling}, they have studied subsets of tasks and languages. 
    We empirically investigate sampling and selection inference strategies, including key methods like Minimum Bayes Risk (MBR) and Best-of-N (BoN)~\citep{huang2025best,eikema-aziz-2022-sampling}, under our massively multi-task constraints, spanning two multilingual LLMs, seven languages and three generative tasks (open-ended generation, math reasoning, machine translation). This broader view lets us pinpoint where existing methods fail and leads us to design methods tailored to generalization across tasks and languages.

    \item \textbf{Novel risk-reducing sampling.} We introduce a novel \textbf{hedged sampling} method to better exploit the sample diversity obtained through sampling under increased softmax temperature. It specifically benefits languages that come with a higher risk of sample quality deterioration at high temperatures. Opposed to prior works that recommend either stochastic or deterministic sampling depending on the task~\citep{song-etal-2025-good}---we show that \emph{including both in the sample set is key to multilingual generalization}. For example, hedged sampling for MBR yields gains of \textbf{+8.2} percentage points on average (across languages and models) increase in win-rates over greedy outputs on mArenaHard-V2; outperforming traditional temperature sampling by +2.2 percentage points.
    \item \textbf{Improved selection strategies.} This leads us to proposing two new selection strategies which capitalize on \emph{long-context modeling and the versatility of cross-lingual generation abilities} of recent generalist LLMs. We call these \textbf{Checklisted One-Pass Selection (CHOPS)} and \textbf{Cross-lingual MBR (X-MBR)} and show these---combined with hedge sampling--- obtain \textbf{+17.3} (Aya Expanse 8B) and \textbf{+9.4} (Qwen3 8B) percentage points average increase in multilingual win rates when evaluated against the single sample baseline. When evaluated in more realistic adverse settings, our methods increase the win-rates on average by \textbf{+6.8} points against the much larger and stronger \textsc{Gemini 2.0 Flash}. Finally, we demonstrate the robustness of these techniques at scale as we improve the performance of \textsc{Command-A}  by an average of \textbf{+9.0} points against single sample decoding. Notably, using the best open source specialized reward model (RM) with BoN scoring only gets a modest +4.5 increase from the same set of samples.
\end{enumerate}

We distill our findings into a \textit{recipe for squeezing the most out of multiple samples in a multilingual and multifaceted generation paradigm}, which we coin the ``Multilingual LLMonade Recipe''.
In our test bed, the recipe for instance improves the Aya Expanse 8B model by up to +17.3 percentage points in multilingual win rates on open-ended generations, +7.9 points accuracy on MGSM, and +0.72 points in XCOMET on WMT24++.\footnote{Which is a notable improvement with respect to the XComet-XL metric, see \url{https://kocmitom.github.io/MT-Thresholds/} \citep{kocmi-etal-2024-navigating}.} This demonstrates how much can be done with as little as 4 more samples without any compute going into training.

Our findings have implications for the broader test-time scaling landscape, as they demonstrate that careful design of sampling and selection techniques can bring important gains even at the low-end scale of inference-time scaling for high-end multilingual LLMs. Contrary to the trend of exploiting specialized reward models for single-task inference-time scaling, we show that even in a challenging and diverse multi-task setup, robust improvements can be found with generalist LLM judges across the bench, leveraging their versatility and on-the-fly adaptability. 

In the following, we will take apart the question on how to optimize multi-sample inference, by first investigating the \textit{sampling strategy} (\cref{sec:sampling}), then comparing multiple \textit{selection strategies} (\cref{sec:selecting}).
Throughout the paper, we mark our newly introduced methods and their empirical effects with \texttwemoji{lemon}, and contrast English results (\english) with non-English (\nonenglish) results.

\section{How to Sample? 
}\label{sec:sampling}

The first ingredient for successful test-time scaling is the creation of a sample pool of sufficient quality: At least one sample in the pool needs to be of higher quality than what can be expected from a single sample. Our research question is here: \emph{How to create a sample pool with a strategy that is robust across languages and tasks?}

\begin{sectionfindings}
\item Even with as little as $N=5$ samples, parallel scaling can improve multilingual generation quality significantly, when done right.
\item Addressing multiple languages and tasks at once with the same inference strategy requires careful design of the sampling algorithm to anticipate higher variance for non-English languages.
\item Adding deterministic outputs, here from greedy decoding, to the sample set for parallel scaling effectively helps hedging risks of subpar sample quality for non-English languages.
\end{sectionfindings}

\subsection{Methodology}
\textbf{Temperature Sampling} Our first task is to create a valuable pool of generations via stochastic sampling. We explore different variants of temperature sampling~\citep{ACKLEY1985147} as it offers an intuitive way of steering the diversity and quality of the generation pool.
Temperature sampling divides the logits $l_{\theta}$ for each token prediction by a fixed constant $\tau > 0$
before the softmax normalization:
\begin{align}
    p^{\tau}_{\theta}(y_t\mid x, y_{<t}) = \texttt{softmax}(l_{\theta} / \tau).
    \label{eq:temp}
\end{align}
Under high temperatures ($\tau>1$) the resulting probability distribution becomes more uniform, while at its extreme (close to 0), it becomes more unimodal.  As a consequence, higher temperatures generate a more diverse pool of samples. The quality, however, varies depending on the task, language, and model. At 0, sampling becomes \emph{greedy} decoding, picking the token with the maximum likelihood at each decoding step, which we refer to as $\tau=0$ for simplicity. 

\textbf{\texttwemoji{lemon}~Multi-Temperature Sampling} Prior works investigated sampling the entire pool from the same temperature~\citep{du2025optimizing, renze-2024-effect, song-etal-2025-good}. When we have a large variety of tasks and languages, representing very different subspaces of the data distribution that the model is trained on, it is impossible to set the temperature optimally for all. To increase robustness, we thus investigate composing the sample set from outputs generated under multiple temperatures $\tau \in [0,1]$. 
The temperatures to sample from can either be chosen according to insights from a development set, or randomly chosen from a range if there is no prior intuition. 
As we will show below, temperature sensitivity across languages and tasks varies, so if temperatures are not tuned individually per task and language---facing combinatorial explosion, especially with massively multilingual and multitask goals---this is better than choosing just one temperature for all settings.

\textbf{\texttwemoji{lemon} Hedged Sampling} As a more controlled variation of multi-temperature sampling, we propose a \emph{hedged sampling} strategy which additionally includes deterministic outputs from greedy search ($\tau=0$) in this mix. This helps hedge risk because even if there are lower-quality samples due to variance caused by higher temperatures, we can default back to strong single-sample candidates for aligned models~\citep{song-etal-2025-good}. Hedged sampling is complementary to token-level techniques that truncate the output space of individual  stochastic samples, e.g. by pruning low-probability tokens with a fixed threshold ($\epsilon$-sampling)~\citep{hewitt-etal-2022-truncation, freitag-etal-2023-epsilon}, or a dynamic threshold based on the model's confidence (min-$p$ sampling) ~\citep{minh2025turning}.
Hedging risks is particularly important for multilingual applications, where the risk tends to be higher in less dominant languages, as we will show in the following experiments.

\begin{figure}
    \centering
    \includegraphics[width=0.8\linewidth]{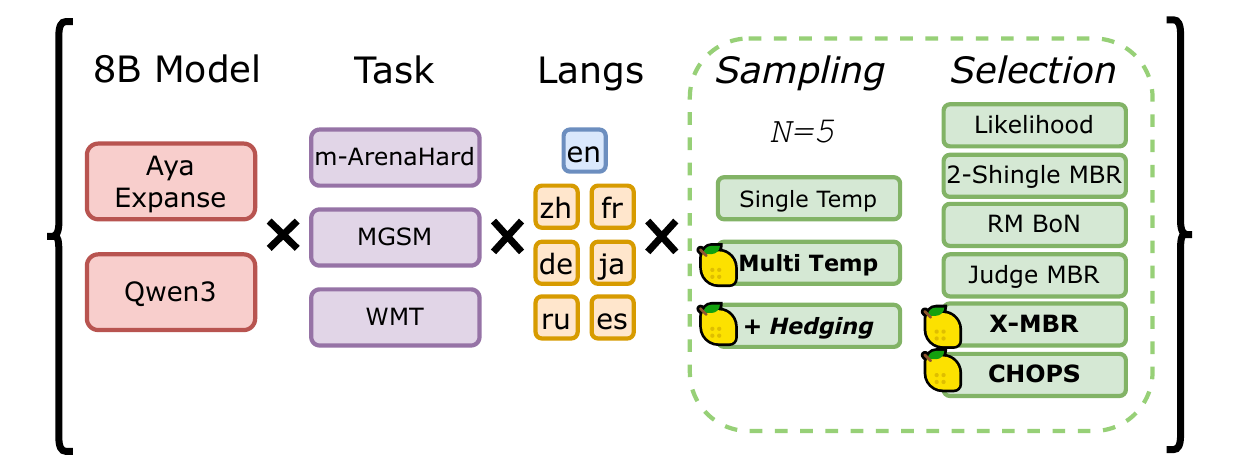}
    \caption{Overview of the \textbf{multilingual multi-task} experimental scope. New methods are marked with \texttwemoji{lemon}.}
    \label{fig:overview}
\end{figure}

\subsection{Experimental Setup}
\textbf{Multilingual Multi-tasking: Models and Benchmarks} 
Open-ended generation tasks have received less attention in test-time scaling works. It is harder to fit a single method or reward model to the diverse challenges that open-ended generations pose. Our goal here is to take a wider view, which means considering both open-ended tasks and tasks with underlying correctness.
The experimental setup is summarized in \cref{fig:overview}. We test \textsc{Aya-Expanse-8B}~\citep{dang2024aya} (short: \textsc{Aya}) and \textsc{Qwen3-8B}~\citep{yang2025qwen3technicalreport} (short: \textsc{Qwen}) models on 7 languages (English, French, German, Japanese, Modern Chinese, Russian, Spanish), selected for their inclusion in the benchmarks for generative tasks that we focus on.
We evaluate on the following benchmarks, detailed in \cref{tab:benchmarks}:

\begin{table}[]
    \centering
    \resizebox{\textwidth}{!}{
    \begin{tabular}{lcccl}
    \toprule
        \textbf{Name} & \multicolumn{3}{c}{\textbf{Data Splits (\# Prompts per Language)}} & \textbf{Metric} \\
        & \texttt{dev} & \texttt{devtest} & \texttt{test} & \\
        \midrule
         Arena & \multicolumn{2}{c}{m-ArenaHard (250/250)} & m-ArenaHard-v2.0 (498) & Win rate\\
        MGSM & \multicolumn{2}{c}{GSM8K-instruct-parallel (250/250)}  & MGSM (258) & Accuracy\\
        WMT & WMT24/15 dev (997/1.5k) & NTREX (1997) & WMT24++ (960) & XComet-XL\\
    \bottomrule
    \end{tabular}
    }
    \caption{Overview of the \textbf{benchmarks} used in this work for open-ended generation (Arena), mathematical reasoning (MGSM), and machine translation (WMT). We compile \texttt{dev} and \texttt{devtest} splits to prevent overfitting our sampling and selecting methods to the \texttt{test} set. For WMT dev, French prompts were retrieved from WMT15 dev, the remaining ones from WMT24 dev. For WMT24++ with originally 998 instances, we skip those marked as ``bad source''.}
    \label{tab:benchmarks}
\end{table}

\begin{itemize}[leftmargin=10pt]
    \item \textbf{Open-ended generation quality: Arena}. We source data from 
    m-ArenaHard~\citep{dang2024aya}\footnote{\url{https://huggingface.co/datasets/CohereLabs/m-ArenaHard}}, an automatically translated version of the English Arena-Hard-Auto v0.1~\citep{li2024crowdsourced} prompts that are diverse and challenging open-ended prompts from LMArena,\footnote{\url{https://lmarena.ai/}} and m-ArenaHard v2.0, 
    created by translating the English portion of Arena-Hard-Auto v2.0\footnote{\url{https://huggingface.co/datasets/lmarena-ai/arena-hard-auto}} with an in-house translation model. 
    We measure win rate \% using GPT-4o (\texttt{gpt-4o-2024-05-13}) as a judge, which is the standard choice for the m-ArenaHard benchmark. Win rates are computed in comparison to (1) greedy decoding for intrinsic comparison and (2) Gemini (\texttt{gemini-2.0-flash}) outputs for extrinsic comparison to a stronger commercial model.
    
    \item \textbf{Mathematical reasoning: MGSM}. 
    Data stems from a collection of translated English grade-school math problems GSM8K~\citep{cobbe2021training}, either translated automatically for 
    GSM8K-instruct-parallel\footnote{\url{https://huggingface.co/datasets/nthakur/GSM8KInstruct-Parallel-instruct-dpo-v0.1}}, 
    or manually for MGSM~\citep{shi2022language}. Final answers are extracted from step-by-step solutions and evaluated for accuracy using exact match, following \texttt{simple-evals}.\footnote{\url{https://github.com/openai/simple-evals/tree/main}}
    
    \item \textbf{Machine translation: WMT.} We leverage several collections of multi-way parallel translations from the yearly shared task on general MT hosted by the Conference of Machine Translation (WMT) ~\citep{federmann-etal-2022-ntrex, deutsch2025wmt24++}. We are only interested in translations from English into other languages, as this requires more complex target language generations.: 
    Translation quality is measured by comparing model outputs against reference translations using the state-of-the-art evaluation metric XComet-XL~\citep{colombo2023xcomet}.
\end{itemize}
This set of generative tasks vary notably in their difficulty and metric sensitivity, so they challenge the robustness of sampling and selection methods that might have been reported successful in prior work on individual tasks.
To ensure rigorous evaluation, we use \texttt{dev} and \texttt{devtest} splits for development of our methods, and only test on \texttt{test} splits once to prevent overfitting.
Full experimental details and the process for creating the exact splits are described in \cref{sec:experiment_details}. Throughout the paper, we refer to these tasks as Arena, MGSM, and WMT respectively, explicitly noting when we are using \texttt{test} splits (\Cref{sec:test}). All other results in \Cref{sec:sampling,sec:selecting} are on the \texttt{dev} splits unless stated otherwise.

\begin{figure}
    \centering
    \includegraphics[width=0.8\linewidth]{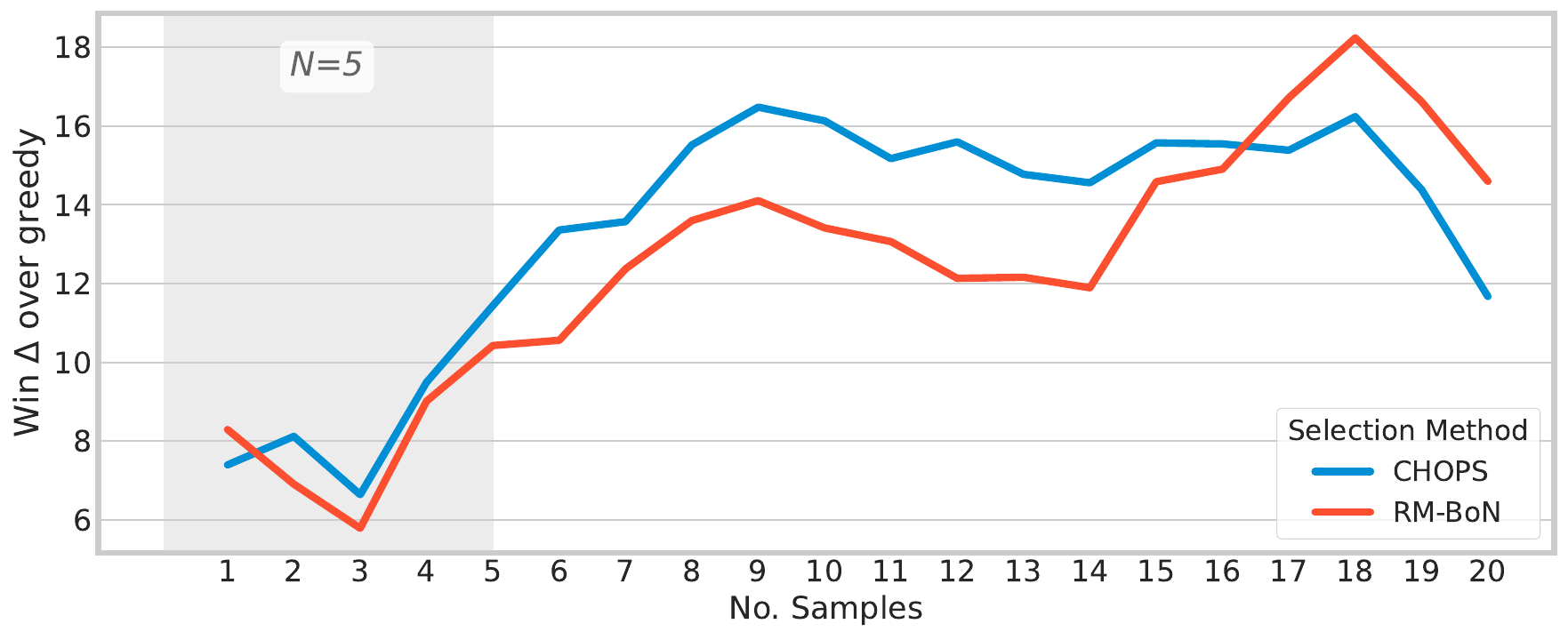}
\caption{\textbf{Multilingual win-rates gains} vs greedy output on the dev set of m-ArenaHard as we increase the sample size from 1 to 20. Performance improvements are steepest at low sample sizes (3-5) with more modest changes beyond that for both selection methods. Results shown for Aya-8B across French, Japanese, and Russian.}
\label{fig:scaling_winrates}
\end{figure}

\textbf{Budget Size for Parallel Scaling}
Compared to prior works that investigate sample sizes in the hundreds to thousands~\citep{freitag-etal-2023-epsilon, song-etal-2025-good, huang2025best}, we focus on the lower end of inference-compute scale. We set $N=5$, given that it is a more realistic workload for large scale production systems (i.e. many inputs applied with 5$\times$ the amount of the normal compute). Complementing this view, scaling curves tend to have their steepest incline in the first steps, i.e. the highest return for additional invested compute, especially for imperfect selection methods~\citep{brown2024large, chen2025we}. In preliminary explorations on open-ended generations, we found that  a perfect match between utility metric and evaluation metric leads to continuous gains with a steep initial increase (see \cref{sec:sample-scaling}). However when the evaluation metric is win rates across languages, returns of scaling are not as guaranteed, in that an even larger inference budget, especially beyond $N=10$, does not necessarily translate to much higher performance as shown in \cref{fig:scaling_winrates}.

\textbf{Measuring Sample Pool Quality}
With an optimistic perspective on the selection methods to work with samples downstream, our main focus lies on the quality of the \emph{best} of the $N$ samples as determined by our evaluation metric $r$, expressing an upper bound of the performance of any selection strategy. For open-ended generations, we query an in-house multilingual reward model to estimate the quality of individual samples, while for the other tasks we can compare the evaluation metric to the reference response. 
In order to have scores that can be compared across tasks and languages, we introduce \textsc{hope} and \textsc{risk}, which intuitively represent the hope and risk of scaling up compared to a single sample. \textsc{hope} is defined as the relative change between the score of the best sample in the set $y^+=\arg \max_{y\in Y}r(y)$ and the evaluation score of the greedy output $\hat{y}$: 
\begin{align}
     \frac{r(y^+)-r(\hat{y})}{r(\hat{y})}.
\end{align}

For \textsc{risk}, we analogously compare the relative change in evaluation score between the worst sample $y^-$ and the greedy output:
\begin{align}
     \frac{r(y^-)-r(\hat{y})}{r(\hat{y})}.
\end{align}
We report \textsc{hope} and \textsc{risk} averaged across instances from each benchmark.

\subsection{Results}

\begin{figure}[h]
    \centering
    \includegraphics[width=0.9\linewidth]{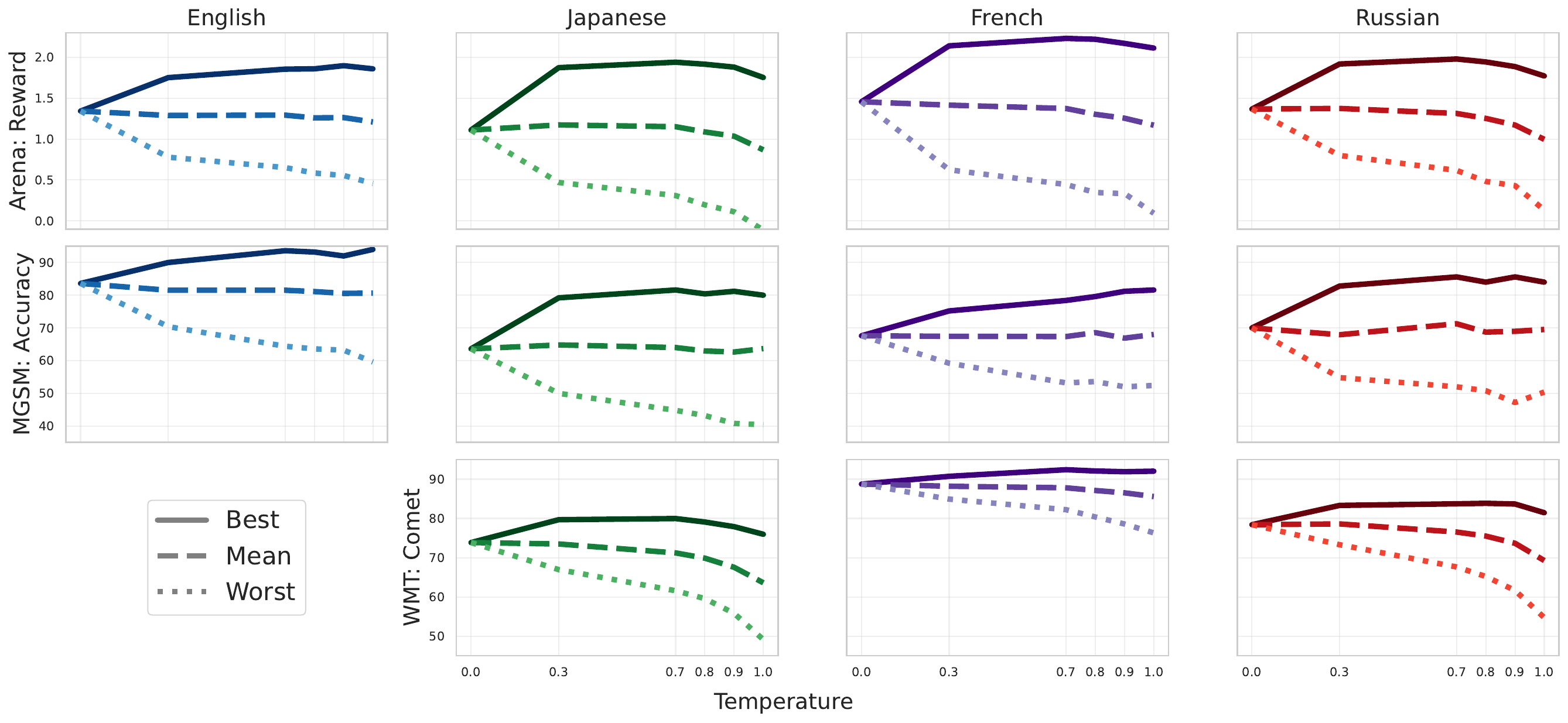}
    \caption{\textbf{Quality under single temperature sampling} For each temperature, we evaluate best, worst and mean quality for $N=5$ samples from Aya for English, Japanese, French and Russian on each of the dev sets of the tasks (rows: Arena, MGSM, WMT). While best-case scores improve over greedy outputs ($\tau=0$), the effect varies across languages and tasks, with a notable drop in worst-case quality for Japanese and Russian Arena and WMT at high temperatures. }
    \label{fig:temperature}
\end{figure}

\textbf{Higher temperature sensitivity for non-English.} In classic single temperature sampling, we dedicate our entire inference budget to sampling at one fixed temperature. \Cref{fig:temperature} compares how best, worst and average performance differ across all three tasks if we spend this budget at different temperatures.
We observe consistent trends: As temperature increases, the gap between best-case and worst-case outcomes widens. While higher temperatures lead to improved best-case scenarios, they also increase the chances of generating lower-quality examples. Notably, the rate at which variance increases is influenced by both the language and the nature of the task. Comparing English against the other languages, its average sample quality is more stable even at higher temperatures (more \emph{eurythermal}), while it drops earlier for other languages. Likewise, best-sample quality continues to grow till close to $\tau=1$ for English (French for WMT), while it starts decaying much earlier and steeper for other languages (here Japanese \& Russian). Trends hold beyond the selection discussed here, see \cref{app:temperatures}.

\textbf{Greedy outputs are the best single-sample bet.} When inspecting the average scores at different temperatures, i.e. their expected quality, it becomes clear that greedy outputs ($\tau=0$) have equal or greater than the \emph{average quality} of outputs at all temperatures, since they are not negatively affected by variance. This is consistent across languages and tasks and corroborates the recommendation for greedy decoding by~\citep{song-etal-2025-good}. Hence, we will compare to greedy decoding as a single-sample baseline when measuring the benefits of scaling up.

\begin{figure}[t]
    \centering
    \begin{subfigure}{0.56\textwidth}
        \centering
        \includegraphics[width=\linewidth]{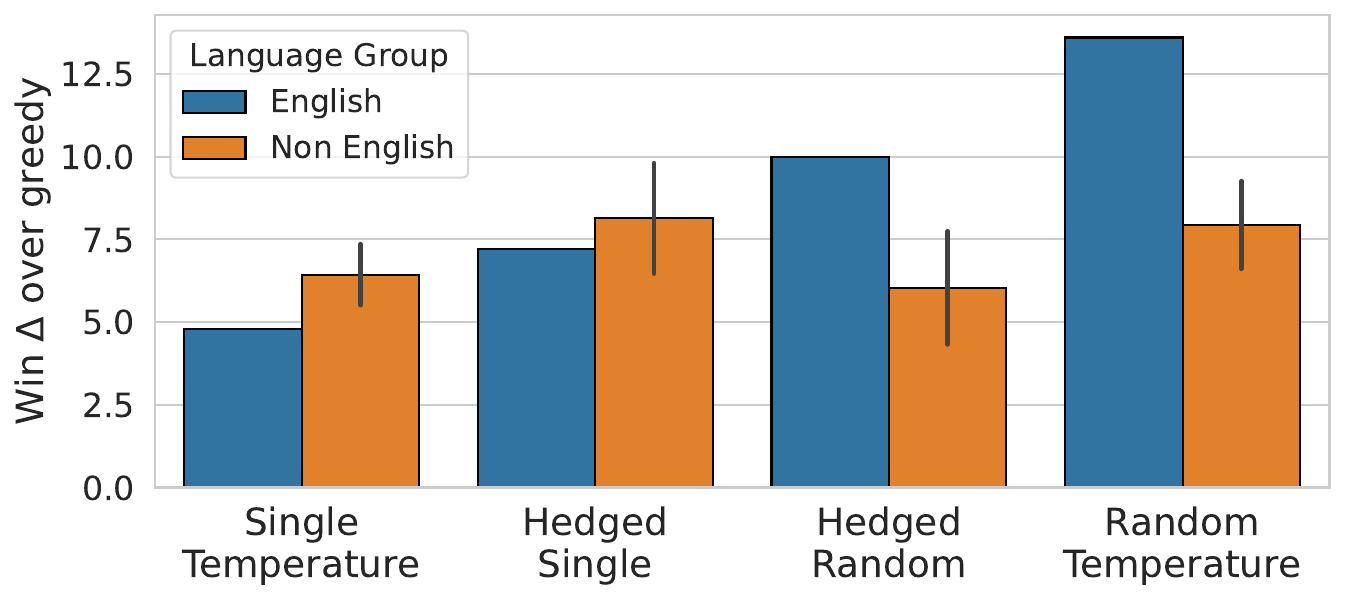}
        \subcaption{Sampling methods comparison}
        \label{fig:sampling_ablation}
    \end{subfigure}
    \hfill
    \begin{subfigure}{0.40\textwidth}
        \centering
        \resizebox{\columnwidth}{!}{
        \begin{tabular}{llrr}
        \toprule
        \textbf{Setting} & \textbf{Task} & \textbf{\textsc{hope}} & \textbf{\textsc{risk }}\\
        \midrule
        \english\,English  & Arena & 38.8 & -51.5 \\
         & MGSM & 11.9 & -23.8 \\
         & \textbf{Average} & \textbf{25.4} & \textbf{-37.7} \\
        \midrule
        \nonenglish\,Non-English  & Arena & 55.5 & -65.3 \\
         & MGSM & 18.7 & -23.7 \\
          & \textbf{Average} & \textbf{37.2} & \textbf{-44.5} \\
        \bottomrule
        \end{tabular}
        }
        \subcaption{\textsc{Risk} \& \textsc{Hope} across tasks and languages.}
        \label{tab:hope_risk_0.7}
    \end{subfigure}
    \caption{Performance analysis of \textbf{temperature sampling methods}. \textbf{Left}: Sampling from a one vs multiple temperatures, with and without hedging: Win rates over greedy outputs on mArenaHard with Judge MBR and $N=5$ samples, averaged across models. We choose $\tau=0.7$ as single temperature, and random temperatures are uniformly sampled from ($\tau \in [0.0, 0.3, 0.7, 0.8, 0.9, 1.0]$). Hedging replaces one sample from the pool with the greedy output. \textbf{Right}: \textsc{hope} and \textsc{risk} of sampling at $\tau=0.7$, relative to evaluation scores of greedy decoding for Aya Expanse 8B. Values are reported as percentages.}

    \label{fig:main_figure}
\end{figure}

\textbf{Higher risk and hope for non-English.} \Cref{tab:hope_risk_0.7} quantifies this trade-off at $\tau=0.7$ for m-ArenaHard and MGSM.
English (\english) sampling yields an average \textsc{hope} of 25.4\% and \textsc{risk} of –37.7\%, whereas non-English (\nonenglish) attains higher \textsc{hope} (37.2\%) but at greater \textsc{risk} (–44.5\%). This illustrates the importance of balancing potential gains against possible losses, particularly in multilingual settings where it is thus more risky to sample at extreme temperatures.
These results also highlight the significant headroom achievable with just five samples.

\textbf{\texttwemoji{lemon} Benefits of multi-temperature sampling} 
With the knowledge of these trends across temperatures, one could pick a single temperature for each task and language to optimize the respective metrics. However, this analysis is expensive since it has to be optimized for each condition separately. If we assume no prior intuition for the task and language, we could instead use our budget to sample across multiple temperatures, ($\tau \in [0.0, 0.3, 0.7, 0.8, 0.9, 1.0]$). We compare this against sampling from a single temperature (here $\tau=0.7$, chosen for its overall highest best-case performance across tasks, languages and models). How does this choice affect downstream performance?
We utilize Judge MBR as a standard selection technique (details will follow in \cref{sec:selecting}) and measure m-ArenaHard win rates for all sampling techniques against greedy outputs, averaging results across our 8B models, \textsc{Aya} and \textsc{Qwen}. \cref{fig:sampling_ablation} shows clear downstream benefits for random multi-temperature sampling over single-temperature sampling, with gains of +8.8 points for English (\english) and +1.5 points for non-English (\nonenglish). We attribute English's larger improvement to its comparatively lower \textsc{risk} at high temperatures (\cref{fig:temperature}) combined with high \textsc{hope}.

\textbf{\texttwemoji{lemon} Hedged sampling reduces risk by including greedy outputs} Given our focus on multilingual generations, we explore approaches to manage the higher \textsc{risk} across languages at high temperatures. To mitigate this risk, we propose \emph{``hedged''} variants, where one sample in the pool is obtained from greedy decoding. 
\Cref{fig:sampling_ablation} visualizes the effect of hedging for m-ArenaHard win-rates over greedy outputs. When we combine 4 samples at $\tau=0.7$ plus the greedy output for the MBR method to choose from, we find that this safety net effectively increases win-rate gains by +2.4 points for English (\english) and provides the largest improvement for non-English (\nonenglish) with +2.2 points.
This constitutes the largest improvement over single temperature sampling in non-English languages.
Our analysis shows that MBR selects greedy for 35.3\% of prompts on average across languages and tasks. 
Overall, single temperature sampling with hedging balances English and non-English performance best, so we choose it as a base for our selection experiments.

\textbf{Combination with probability pruning} Furthermore, hedged sampling can be combined with techniques that reduce risk at the token level, such as min-$p$~\citep{minh2025turning}. In our setup, we find that for the majority of tasks and selection methods configurations, min-$p$ provides additional gains over hedged sampling alone. The improvements are most consistent for machine translation, where similar probability pruning techniques have previously been shown to be essential for  MBR~\citep{freitag-etal-2023-epsilon}. The results in \Cref{sec:min_p_ablations} confirm these gains across tasks and selection methods. Therefore, we incorporate min-$p$ into our hedged sampling approach for test set evaluations and denote this combination with the subscript min-$p$.

This discussion highlights how even simple adjustments to the composition of a small sampling pool can have noticeable downstream effects. Before turning to optimizing selection methods, we summarize our recommendations for squeezing the most of temperature sampling:

    \begin{center}
        
\begin{tcolorbox}[
    title=Multilingual LLMonade Recipe \textbf{Part I},
    colback=yellow!10, 
    colframe=yellow!50,
    coltitle=black, 
    width=0.9\textwidth
    ]

    \textbf{Step 1:} Use hedged single-temperature sampling at a moderate temperature (0.7--0.9) to generate $N$ samples. \\
    
    \textbf{Optional:} Localize a reasonably high temperature for different contexts. 
\end{tcolorbox}
\end{center}

\section{How to Select?}
\label{sec:selecting}
Once we have sampled a pool of generations with promising quality, our goal is to correctly identify the best generation in our pool of candidates. Our research question here is: \emph{How to select from a sample pool with a strategy that is robust across languages and tasks?}

\begin{sectionfindings}
    \item Strong reward models or judges are required to robustly select the best of the sampled candidates. Our methods based on a multilingual generalist LLM as judge are competitive with BoN selection with a specialized RM.
    \item Crosslingual evidence (our proposed X-MBR) can further boost improvements over single-sample decoding.
    \item One-pass selection (our proposed CHOPS) is a simple and cost-effective solution especially for multilingual open-ended tasks.
    \item X-MBR and CHOPS can even improve the strong judge LLM itself (here Command A), confirming their robustness and competitiveness.
\end{sectionfindings}

\subsection{Methodology}
In the following section, we briefly review multiple selection techniques of varying complexity, and propose our own extensions (\texttwemoji{lemon}) that are particularly equipped for multilingual generative tasks.

\textbf{Maximum Likelihood} Given a pool of samples \( Y \),
the sample \( \hat{y} \) with the highest likelihood under the model distribution \( p_{\theta}(y\mid x) \) should be a good candidate for selection when the model is well calibrated (i.e. likelihood and quality correlate): 
\begin{align}
\hat{y} = \arg\max_{y \in Y} p_{\theta}(y\mid x).
\end{align}
This constitutes an intrinsic metric that relies only on the sampling model itself.

\textbf{Best-of-N (BoN)} introduces an \emph{extrinsic utility metric} \( U(y) \) 
to score each sample independently. The selected sample \(\ \hat{y} \) is then the one with highest utility score: 
\begin{align}
    \hat{y} = \arg\max_{y \in Y}U(y).
\end{align}
This approach relies on the utility metric being well aligned with the task evaluation metric and well calibrated, i.e. rating outputs adequately on a common scale even if scored independently. 
It is commonly used with specialized reward models (RM BoN)~\citep{zhang2024generative,ichihara2025evaluation,pombal2025m, son2025linguistic} or verifiers in math or code tasks~\citep{snell2025scaling, cobbe2021training, lightman2024lets, zhang2025generative}. 
One could also leverage an LLM for absolute BoN scoring, but this approach was not competitive out-of-the-box in preliminary explorations, as it rated all outputs similarly high.

\textbf{Minimum Bayes Risk (MBR)} decoding searches for the candidate \( \hat{y} \) that \emph{minimizes the expected risk} over the distribution of samples~\citep{kumar-mbr,kumar-byrne-2004-minimum,eikema-aziz-2020-map,eikema-aziz-2022-sampling}. The risk \( R(y') \) of a candidate \( y' \) is approximated with pairwise comparisons from a sample pool $Y$: 
\begin{align}
R(y') \approx \frac{1}{|Y|} \sum_{y \in Y} L(y, y'),
\end{align}
where \( L(y, y') \) is a pairwise loss function measuring the discrepancy between a candidate \( y \) and and a pseudo-reference \( y' \). 
MBR thus selects 
\begin{align}
    \hat{y} = \arg\min_{y' \in Y_h} \sum_{y \in Y_e} L(y, y'),
    \label{eq:mbr}
\end{align}
where \( Y_h \subseteq Y \) is the hypothesis set, and \( Y_e \subseteq Y \) is the evidence set used to estimate the risk. As \citet{bertsch-etal-2023-mbr} highlighted, the evidence set \( Y_e \) aims to cover a representative portion of the space for accurate risk estimation, while the hypothesis set \( Y_h \) focuses on the narrower, high-quality region to avoid considering low-quality candidates, but they do not need to be identical.

For loss functions, there are many possible implementations: When aligned well with the task evaluation metric, it \emph{reduces the optimization-evaluation gap} and brings larger empirical gains~\citep{kovacs-etal-2024-mitigating}, which has made it a popular method in machine translation and open-ended generation~\citep{fernandes-etal-2022-quality,freitag-etal-2023-epsilon, wu2025better}.
In this way, we can optimize for pairwise comparisons under an LLM judge at test time, such as win-rate evaluations. When loss functions focus on similarity (e.g. token-based similarity, which we will test with 2-shingles), MBR becomes the equivalent to majority voting in classification tasks~\citep{bertsch-etal-2023-mbr}. It selects the sample that is most \emph{consistent} with the evidence set, which relates it to the notion of self-consistency~\citep{wang2023selfconsistency,shi-etal-2024-thorough,chen2025we,wang2025thinkdeepthinkfast}.

\textbf{\texttwemoji{lemon}~Checklisted One-Pass Selection (CHOPS)} 
Most of the prior selection methods present considerable computational cost: BoN requires \(\mathcal{O}(N)\) forward passes, and MBR even \(\mathcal{O}(N^2)\) due to pairwise comparisons. This may be a reasonable approach for some latency-insensitive tasks, but we explore proposing an alternate approach that reduces this efficiency penalty. Capitalizing on the development of longer context windows for LLMs and their efficient processing,  we prompt a judge model to first generate a checklist tailored to the given task, then evaluate all candidate samples against this checklist to directly \emph{choose the best response in one pass} (see prompt in \cref{app:prompts}). This is inspired by the success of rubrics to facilitate LLM judge decisions~\citep{kim-etal-2024-prometheus} and the ability of LLMs to generate prompt-specific checklists~\citep{cook2024ticking}, which might help to adapt the judge on-the-fly to diverse selection scenarios across languages and tasks. 
CHOPS requires only \(\mathcal{O}(1)\) forward pass of the LLM judge, fitting all samples into the LLM context in a single call.
While BoN rates samples independently, CHOPS rates them in one global context, anchored in the checklist criteria, which arguably alleviates concerns of LLM judge calibration across multiple independent ratings. 
Our ablation reported in \cref{sec:chops_ablation} confirms the effectiveness of the self-generated checklists in contrast to a more straightforward prompt, especially for non-verifiable tasks and languages other than English. 

\textbf{\texttwemoji{lemon}~Crosslingual MBR (X-MBR)}\label{sec:xmbr} We propose X-MBR as a method to improve performance by utilizing the multilingual capabilities of LLM models. Building directly on the MBR paradigm, X-MBR uses cross-lingual evidence to more robustly select from target-language candidates. For an input \(x\), X-MBR uses the same \textit{hypothesis set} as standard MBR, i.e. the same five samples in the target language.
The novelty lies in the \textit{cross-lingual evidence set} $Y_{ex}$, that extends original in-language evidence set $Y_e$ from \cref{eq:mbr} by a smaller set of cross-lingual samples (\(M < N\)). These samples are generated by instructing the same LLM to respond in a different ``evidence'' language
(e.g.\ English, see prompt in \Cref{app:prompts}). This approach does not require prompt translation, it directly prompts the multilingual model to output the response in another language.
We then pick the candidate from the hypothesis set that accumulates the highest cross-lingual support, with the same MBR selection criterion as for classic MBR (\cref{eq:mbr}), but including additional cross-lingual comparisons by the LLM judge between the hypothesis set $Y_h$ and the set of new crosslingual samples $Y_{ex}$: 
\begin{align}
    \hat{y} = \arg\min_{y' \in Y_h} \sum_{y \in (Y_e \cup Y_{ex})} L(y, y').
    \label{eq:xmbr}
\end{align}
This exploits both the cross-lingual generation abilities of the model that we sample from, as well as the cross-lingual comparison abilities of the judge LLM. X-MBR requires generating $N + M$ total samples and performing $N \times (N + M)$ pairwise comparisons, approximating $\mathcal{O}(N^2)$ complexity. We set $M = 3$ cross-lingual evidence samples, which increases the compute budget but in turn it improves precision of the selection, as we will see in the next section. It is an interesting direction which explores the return on additional compute investment through cross-lingual validation.

\subsection{Experimental Setup}

\begin{figure}[t]
    \centering
    \includegraphics[width=0.7\linewidth]{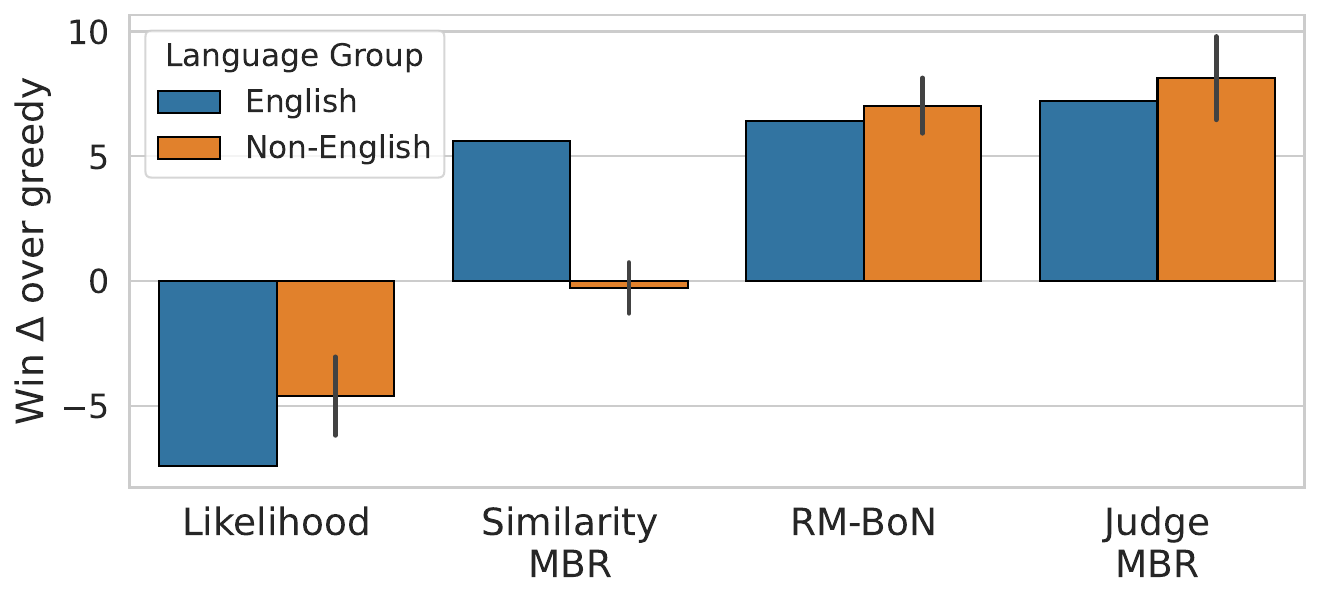} 
    \caption{Comparison of \textbf{baselines vs RM and LLM Judge} on $N=5$ generations in terms of the win-rate comparing to greedy outputs on mArenaHard. Averaged across models, and non-English languages.}
    \label{fig:baselines}
\end{figure}

\textbf{Baseline} We compare against \textit{greedy} decoding, which in each generation step selects the token with the highest model probability, resulting in deterministic and predictable output. This gives us a simple yet effective comparison point~\citep{song-etal-2025-good}, as we have empirically confirmed in ~\cref{sec:sampling}. In this way we can quantify the benefits from scaling the generation budget from 1 to 5 samples. For open-ended generation, we additionally establish a more challenging benchmark by comparing selected samples from our models against single greedy samples from the significantly larger Gemini 2.0 Flash model, ensuring a more rigorous assessment.

\textbf{Choice of Utility Metrics}
Some of the above selection methods can be used with multiple utility metrics or backbone models, such as BoN or MBR. In order to disentangle the effect of the method from the underlying utility metric, we compare multiple instantiations of each. 
Concretely, we benchmark two versions of MBR, \textbf{2-Shingle MBR}---which relies on the simple Jaccard similarity of token-level 2-grams from pairs of generations, and \textbf{Judge MBR}---which queries an LLM judge for pairwise comparisons. 
In preliminary experiments, we also explored using the LLM judge for BoN vs the RM, but the LLM judge did not perform competitively due to missing calibration for generating absolute scores. 

\textbf{Choice of RM and Judge Model} Based on prior findings that the precision of the utility score can have major impact on the success of inference-scaling \citep{huang2025best, stroebl2024inference}, we aim to pick the best scoring open judge or RM model for our experiments. For techniques which use an LLM judge, we use \texttt{Command A}~\citep{cohere2025command}, an 111B model optimized for multilingual performance supporting 23 languages. Command A scores competitively to GPT-4o on mRewardBench~\citep{gureja2024m}. For BoN RM, we choose the leader from the RewardBench leaderboard~\citep{RewardBench} \texttt{INF-ORM-Llama3.1-70B}~\cite{INF-ORM-Llama3.1-70B}, which is based on the multilingual Llama3.1 base. INF-ORM-LLama3.1-70B is trained on a mix of open-sourced preference pairs~\citep{liu2024skywork} with difference magnitudes determined by GPT-4o for scaling~\citep{wang2024helpsteer2}.
This constitutes a very strong competitor for any LLM judge approach, as this model is specifically engineered and trained to perform generation scoring aligned with GPT-4o. \Cref{app:reward_judge_choice} details the selection process of RM and LLM judge. Each selection method with LLM judge requires a specific prompt, which we list in ~\cref{app:prompts}. For consistency, the instruction prompts are always in English.

\begin{table}[t]
\centering
\resizebox{0.6\columnwidth}{!}{
\begin{tabular}{lccccc}
  \toprule
  \textbf{Model} & \textbf{Lang.} & \textbf{S‐MBR} & \textbf{En‐MBR} & \textbf{Zh‐MBR} & \textbf{R‐MBR} \\
  \midrule
  \multirow{2}{*}{\textsc{Aya}} 
    & \english      & 12.00 & --     & \textbf{15.20} & 13.20 \\
    & \nonenglish  & 13.27 & \textbf{15.93} & 12.64 & 11.53 \\
  \midrule
  \multirow{2}{*}{\textsc{Qwen}} 
    & \english      & 9.20  & --     & \textbf{20.00} & 4.40  \\
    & \nonenglish  & 6.07 & \textbf{9.87}  & 8.40  & 7.53  \\
  \bottomrule
\end{tabular}
}
\caption{\textbf{Expanding MBR evidence sets for X-MBR:} Comparison of different sources of evidence for 3 additional samples: S: from the same language, R: random samples across languages, En: from English, Zh: from Chinese. The benefits of for win-rates on mArena are highest when sampling evidence from more dominant languages.} 
\label{tab:xmbr_ablations}
\end{table}

\subsection{Results} \label{sec:selection_res}
First (\cref{sec:dev}), we compare selection methods that do not rely on LLM judges or RMs (Maximum Likelihood, Similarity MBR) with those that do rely on them (RM BoN, Judge MBR), and develop our newly proposed selection methods (CHOPS, X-MBR). These initial experiments are based on on the development set of m-ArenaHard-v0.1 benchmark and hedged sampling at $\tau=0.7$.
Second (\cref{sec:test}), we put our final best RM/judge-based methods to the test on the held-out test sets across all three tasks of open-ended generation, machine translation and math reasoning.

\subsubsection{Establishing the Best Selection Methods}\label{sec:dev}

\textbf{Maximum Likelihood and Similarity-based MBR are not competitive crosslingually.} From \Cref{fig:baselines}, we observe that Maximum Likelihood selection results in losses across the bench, suggesting that the model's internal probabilities are not well calibrated for win rate evaluation. Averaged across our models, similarity‐based MBR, which selects the sample most similar to others in the pool, yields a \textit{+5.6}\% increase in win-rate over greedy in English (\english ) but provides no improvement in non-English (\nonenglish). We suspect this is due the quality of the pool of samples which does not sustain picking the most consistent sample, as it is expected to have higher variance compared to English (see \cref{sec:sampling}).

\textbf{BoN and judge‐based selection outperform greedy.}
RM BoN shows consistent improvements over greedy decoding for both groups (\english \textit{6.4}\% in English, \nonenglish \textit{7}\% in non‐English), establishing it as a strong baseline. This corroborates recommendations by \citet{wu-etal-2024-reuse} to use reward models crosslingually for BoN when they have a multilingual LLM backbone. 
Judge‐based MBR achieves the highest deltas (\english \textit{7.2}\% in English, \nonenglish \textit{8.1}\% in non‐English), showing that a general-purpose multilingual LLM judge can outperform a specialized reward model. The flexibility of using LLMs for pairwise comparisons in the MBR setup aligns well with the pairwise setup in win rate evaluation, though it requires $O(N^2)$ comparisons at test time.

\textbf{\texttwemoji{lemon} Better judgment with crosslingual evidence}
We extend the evidence set of MBR with $M=3$ more samples, either (1) samples of the same language (S-MBR)---which corresponds a target-language extension of the standard MBR, (2) a hard-coded fixed language (English, Chinese; En/Zh-MBR), or (3) randomly chosen languages for all prompts (R-MBR). For randomly chosen languages we sample uniformly from the set of languages of our experimental setup.
In \cref{tab:xmbr_ablations}, we compare these variants of X-MBR. Note that they all use the same candidates for selection, they just differ in the composition of the evidence set. 
Using additional evidence from the same language results in a an average 10\% win rate across \textsc{Aya} and \textsc{Qwen} over greedy, compared to 7.6\% from MBR without extended evidence. 
Both En-MBR and Zh-MBR result in significant improvement over both greedy (up to 14\% for Zh-MBR) and the S-MBR baseline, particularly in non-English languages (\nonenglish). This shows that it is best to sample evidence from dominant other languages.
This works better than randomly drawing samples from a mix of languages (R-MBR). Hence, we set this as the default for all following experiments with X-MBR: Sample in Chinese, if responding to an English prompt, else sample in English.
The results presented in this experiment demonstrate that we can effectively leverage the model's multilingual capabilities to enhance performance across all languages. Furthermore, this underscores the potential of multilingual LLM judges with strong crosslingual capabilities to optimize available test-time compute. A key ingredient in this part of the recipe is the careful selection of cross-lingual approaches.

\begin{table}[pt]
    \centering
    \begin{tabular}{lllccc|c}
    \toprule
    \textbf{Task} & \textbf{Model} &  & \textbf{RM BoN$_{\text{min-$p$}}$} & \textbf{CHOPS$_{\text{min-$p$}}$} & \textbf{X-MBR$_{\text{min-$p$}}$} & \textbf{\emph{Greedy}} \\
    \midrule
    
    \multirow{4}{*}{\textbf{Arena}} & \multirow{2}{*}{\textsc{Aya}} & \english      & \textbf{19.60} & 14.40 & 16.80 & -- \\
                                                            & & \nonenglish & 16.27 & \textbf{17.33} & 15.67 & -- \\
    \cmidrule(lr){2-6}
    & \multirow{2}{*}{\textsc{Qwen}} & \english        & 2.00 & 7.60   & \textbf{8.80} &  --\\
                             & & \nonenglish    & 5.87         & 8.27 & \textbf{9.40} & -- \\
    \midrule
    \multirow{4}{*}{\textbf{MGSM}} & \multirow{2}{*}{\textsc{Aya}} & \english      & 7.76 & 6.96 & \textbf{7.76} & 77.84 \\
                                                            & & \nonenglish & \textbf{9.59} & 6.19 & 7.92 & 69.55 \\
    \cmidrule(lr){2-6}
    & \multirow{2}{*}{\textsc{Qwen}} & \english        & \textbf{3.04} & 1.84   & 0.64 &  94.96\\
                             & & \nonenglish    & 3.65          & 2.19 & \textbf{3.85} & 84.41 \\
    \midrule
    \multirow{2}{*}{\textbf{WMT}} & \textsc{Aya} & \nonenglish & \textbf{1.04} & 0.72 & 0.20 & 71.92 \\
    & \textsc{Qwen}                              & \nonenglish & \textbf{1.43} & 1.12 & 0.93 & 76.15 \\
    \bottomrule
    \end{tabular}
    
    \caption{\textbf{Test set results}: Quality gains over greedy decoding by selecting from five samples (\texttwemoji{lemon} hedged $\tau=0.7$ and $\text{min-}p$ with $p=0.2$) for 7 languages of study. X*-MBR uses Chinese as evidence languages for English, and English for the rest. Highest values for each row are \textbf{bold}.}
    \label{tab:test_set}
\end{table}

\subsubsection{Testing in Multilingual Multitasking}\label{sec:test}

\textbf{Bringing it all together: Test set performance across tasks and models.} \Cref{tab:test_set} present the final results for our LLM-judge based aggregation methods using hedged sampling with $\tau=0.7$  and min-$p$ with $p=0.2$ across models and tasks. We find improvements over the greedy baseline (i.e., the best single-sample method) in all tasks, languages, and methods with magnitudes of improvement that are substantial; considering that we are working with as few as five samples to choose from. 

\textbf{CHOPS vs X-MBR} For open-ended tasks, both CHOPS and X-MBR outperform reward model-based BoN selection. Across models, \emph{BoN} achieves win deltas of 10.8\% (English \english) and 11.0\% (non-English \nonenglish) over greedy decoding. \emph{CHOPS} improves performance to 11.0\% (\english) and 12.8\% (\nonenglish), showing strength in multilingual settings. \emph{X-MBR} performs best overall with 12.8\% (\english) and 12.5\% (\nonenglish), respectively. In close-ended evaluation, selecting from 5 samples yields significant gains across all methods, even with strong initial greedy performance. BoN often performs best, particularly for WMT. CHOPS overall yields consistent gains with lower compute costs. X-MBR shines particularly for non-English Qwen, where there are larger language disparities to bridge.

\begin{table}[t]
    \centering
    \begin{tabular}{clccc|c}
    \toprule
     &  & \textbf{RM BoN }    & \textbf{CHOPS}  &  \textbf{X-MBR} &       \\
     \textbf{Open-Ended Eval} &  & $O(N)$  & $O(1)$ & $O(N(N+M))$ & \textbf{\emph{Greedy} }   \\
    \midrule
    
    \multirow{2}{*}{ \textsc{Aya} vs Gemini}  & \english     & \textbf{7.60}         & 2.80   & 3.60 & 20.80  \\
                                & \nonenglish  & 6.87         & 5.07   & \textbf{7.33}   & 25.80 \\
                             
    \cmidrule{2-6} 
     \multirow{2}{*}{ \textsc{Qwen} vs Gemini}  & \english     & 1.60         & \textbf{2.80}   & 0.40 & 38.40   \\
                  & \nonenglish  & 6.00         & 5.00   & \textbf{6.33} & 42.07   \\

    \bottomrule
    \end{tabular}

    \caption{\textbf{Open ended test results with extrinsic comparison}: We show gains in win rates in a more challenging setup: against \textit{Gemini 2.0 flash} as a strong baseline. We report the delta in win rates from using 5 samples with a specific selection method compared to the single sample.}
    \label{tab:gemini_cmdA}
    \end{table}

\textbf{LLMs-as-Judge based methods are more robust in open-ended tasks.} We conduct two additional evaluations to assess robustness beyond greedy baseline comparisons. In \Cref{tab:gemini_cmdA}, we compare our 8B models against the much larger and more capable Gemini 2.0 Flash model. We report the delta in win-rates obtained from the selection method with 5 samples compared to a single sample. X-MBR achieves the highest multilingual (\nonenglish) gains with an average of +6.8 points, while BoN and CHOPS follow with +6.4 and +5.0, respectively. 

\begin{figure}[t]
    \centering
    \includegraphics[width=\linewidth]{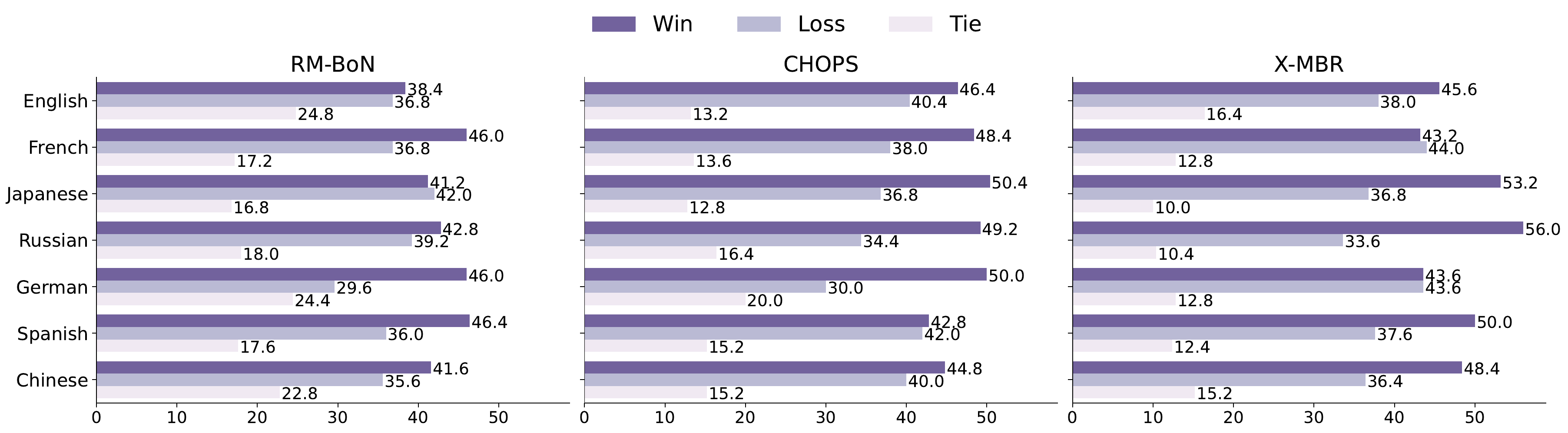} 
    \caption{\textbf{Self-improvement with parallel scaling:} Command A win rates against the greedy single sample baseline for each of the selection methods. Our methods, which also use Command A for selection, outperform the RM BoN in the majority of languages.}
    \label{fig:cmd_a_winrates}
\end{figure}

\textbf{Gains transfer to self-improvement of a high-end LLM.} One could argue that the improvements are only due to performance or size gradients between the smaller 8B models that we have tested, and the much larger RM/LLM judge models. However, improvements can still be found at larger scale and without this gradient:
In \Cref{fig:cmd_a_winrates}, we evaluate a \emph{self-improvement} scenario where we use Command A to both generate samples and perform selection for CHOPS and X-MBR. Not only do we find consistent gains across languages, but also our new selection methods  outperform RM BoN, with CHOPS and X-MBR obtaining remarkable average deltas of 9.0\% and 8.3\% compared to BoN's modest 4.5\%. This consistent cross-lingual performance demonstrates the robustness of our methods even when the same model must evaluate its own outputs.

This leads us to complete the second part of our recommended recipe for parallel scaling of multilingual LLMs in generative tasks:

\begin{center}
\begin{tcolorbox}[
    title=Multilingual LLMonade Recipe \textbf{Part II},
    colback=yellow!10, 
    colframe=yellow!50, 
    coltitle=black,
    width=0.9\textwidth
    ]
    \textbf{Step 2:} Use CHOPS to select the best sample. If you can afford $O(N^2)$ calls to the LLM judge and have a task with strong language disparities, use X-MBR.\\
    
    \textbf{Optional:} A small exploration of multilingual LLM judges to find the best suited one. 
\end{tcolorbox}
\end{center}

\section{Related Work}

\textbf{Stochastic vs Deterministic Inference}
Early LLM research suggested that diversity through stochastic inference often comes 
at a cost of quality~\citep{Holtzman2020The},  benefiting some tasks while hindering others~\citep{Holtzman2020The, kar105743, renze-2024-effect}. 
In the context of multi-sample inference, \citet{song-etal-2025-good} found that the optimal approach depends on the task: closed-ended and verifiable tasks favor deterministic decoding, while open-ended generation benefits from the variance introduced by stochastic sampling.
\citet{du2025optimizing} optimize the temperature across math and coding tasks for English with insights from entropy measures for high samples sizes ($N=256$), but excluding $\tau=0$.
In our work, we focus on exploiting variance in the smaller sample range ($N=5$) across \emph{multiple generative tasks}.
We add the dimension of language that has previously been ignored: \emph{Going beyond English}, higher temperatures pose a higher risk, and when running inference for multiple languages at once, this requires more caution. We address this by combining stochastic and deterministic inference.

\textbf{Multilingual Test-time Scaling and Alignment}
While most test-time scaling and alignment research focuses on English, a few recent works have explored multilinguality. \citet{pombal2025m} propose a multilingual judge LLM for BoN, showing improvements in win rates across three languages.
Similarly, the contemporaneous work by \citet{gupta2025testtimescalingrepeatedsampling} confirms the potential benefits of RM BoN for multilingual open-ended tasks. They find improvements in win-rates for a dozen languages with RM BoN for 5 to 100 samples across model sizes.
 Our study expands on these previous RM BoN explorations with a broader set of tasks and novel methods, specifically crafted for the challenges in multilingual generation both on the sample generation and the sample selection side.
 What emerges as a consistent pattern in their and our work is that RMs appear to generalize well across languages for parallel scaling, even if trained only on rewards for English 
\citep{wu-etal-2024-reuse}.
Similarly, \citet{yong2025crosslingual} demonstrate cross-lingual scaling benefits in math and STEM reasoning with a LLM with a multilingual backbone tuned for English reasoning. They show benefits of non-target language reasoning/scaling, which is loosely related to the effectiveness of crosslingual evidence for X-MBR that we find in our experiments. With the shared motivation to reduce imbalance across languages, \citet{yanglanguage} and \citet{zhu-etal-2024-question} use cross-lingual sample generation with a translation pipeline, while \citet{yoon-etal-2024-langbridge} combine expert models for task and language expertise. Our X-MBR approach achieves significant gains without intermediate translation or experts, directly leveraging the LLM's strong cross-lingual generation capabilities.

\section{Conclusion}
We have conducted extensive experiments on three generative tasks to compile a recipe for multilingual parallel scaling that generalizes across both tasks and models. Based on our insights on the impact of temperature on sample pool quality, we designed a hedged temperature sampling variant, and combine it with selection methods tailored towards multilingual judges. We propose two approaches which improve upon existing methods: Checklisted One-Pass Selection
(CHOPS) and Cross-lingual MBR (X-MBR). These techniques show consistent cross-lingual gains in the benefits of test-time scaling. This has not only implications for inference, but also for applications where multilingual inference is an intermediate step in model improvement, e.g. for synthetic data generation~\citep{thakur-etal-2024-leveraging,dang-etal-2024-rlhf,odumakinde2024multilingual} or distillation~\citep{zhang-etal-2024-enhancing-multilingual}, test-time alignment~\citep{sun2024fast,amini2025variational} or model fine-tuning~\citep{touvron2023llama,snell2025scaling}. Moreover, the gains observed at small scale and the steep incline suggest even greater potential lies at the close horizon when scaling these methods to larger sample pools. Additionally, the effectiveness of using models for both generation and selection opens promising avenues for self-improvement frameworks in multilingual settings.

\section*{Limitations}
\textbf{Reliance on judge alignment} All methods that use extrinsic signals (reward models or LLM judges) for selecting from multiple samples are bounded by their alignment with the evaluation metric, as has previously been pointed out in~\citep{stroebl2024inference,huang2025best}. Our methods do not directly address this issue. By selecting the latest and most generalist judge models for selection, we hope that the effects of task-specific reward hacking / mismatch are reduced.

\textbf{Language selection} Our selected languages are all high-resource languages and well represented throughout the stages of LLM training. Our study does not cover the test case of generalizing to underrepresented languages that are unsupported by the model or not included in stages beyond base model training. We can expect that both quality of samples and LLM aggregation precision will be significantly lower, so approaches like X-MBR that leverage crosslingual knowledge might be more promising.

\textbf{Cost of selection method} We found that a larger generative model was needed to improve upon the base model performance (based on preliminary explorations with mPrometheus~\citep{pombal2025m}) with small $N$. In practice, there is a balance to be found between scaling up $N$ versus scaling up the judge model. Distilling the outputs of the larger generative judge into a smaller model might be an interesting avenue for optimizing that trade-off.

\section{Acknowledgments}
We thank Ahmet Üstün, John Dang, Samuel Cahyawijaya, Arash Ahmadian and other colleagues at Cohere and Cohere Labs for their support and thoughtful feedback. We also thank Sander Land for his contributions to our evaluations.

\bibliography{main,anthology,custom}

\clearpage
\appendix

\clearpage
\appendix

\section{Detailed Experimental Setup}\label{sec:experiment_details}

\textbf{Models and Language Coverage} For the multilingual generative model, we consider two 8B models from different families: Aya-Expanse-8B~\citep{dang2024aya} and Qwen3-8B~\citep{yang2025qwen3technicalreport}. \textsc{Aya-Expanse-8B} is an open-weights multilingual LLM supporting 23 languages. It employs a post-training recipe focused on open-ended generative tasks that includes supervised fine-tuning, multilingual preference tuning~\citep{aakanksha2024multilingual,dang-etal-2024-rlhf} and synthetic data optimization~\citep{odumakinde2024multilingual}, and model merging~\citep{ahmadian2024mix} to achieve strong competitive performance on open-ended benchmarks for models of this size. \textsc{Qwen3-8B} is an open-source dense model from the Qwen3 model family, supporting up to 119 languages and dialects. It was post-trained through distillation from larger models in the family and is used here in its ``non-thinking'' mode without reasoning.  We focus on a subset of 7 languages which are covered by both models: \textit{English, French, German, Japanese, Russian, Simplified Chinese, Spanish} prioritized due to their inclusion in generative benchmarks of interest. For WMT, we translate from English into all other languages.

\textbf{Development Sets}
To ensure the generalization of our methods to unseen data, we follow best practices and meticulously craft a development set and development testing set (\texttt{devtest}) for our evaluations. For all development and test set we use 250 randomly sampled prompts, we describe them below:

\textbf{m-ArenaHard}: We sample 250 prompts from the original Arena-Hard-Auto dataset~\citep{li2024crowdsourced} by randomly picking one from each of the 250 clusters. In addition to this stratification across clusters, we use the translations of the exact prompts from mArena-Hard for each language when creating our development set. This way we ensure no cross-contamination across both languages and clusters.

\textbf{m-ArenaHard-v2.0}:
To create the m-ArenaHard-v2.0 test set, we obtain the 750 prompts from Arena-Hard-v2.0\footnote{\url{https://github.com/lmarena/arena-hard-auto/blob/main/data/arena-hard-v2.0/}} and use papluca/xlm-roberta-base-language-detection\footnote{\url{https://huggingface.co/papluca/xlm-roberta-base-language-detection}} to perform language identification. Of these, the 498 identified as "English" were then translated into 23 languages by using an in-house state-of-the-art translation model. This results in a total test set of 11,454 multilingual prompts.

\textbf{WMT}: We use the development sets from WMT24~\citep{kocmi2024findings} for most language pairs with the except of \texttt{en-fr}, which we obtain from WMT15~\citep{stanojevic2015results}.

\textbf{MGSM}: We obtain the GSM8K Parallel Translated Corpus~\citep{chen2023breaking} and group them by the original prompt. We then randomly select 250 prompt groups and select the same for all languages to avoid cross-lingual contamination. For each math problem, the model is instructed in the specific language to solve step-by-step and provide a final answer. Final answers are extracted from the step-by-step solutions and evaluated for accuracy using exact match to the correct answer, following \texttt{simple-evals}\footnote{\url{https://github.com/openai/simple-evals/tree/main}}.

\textbf{Model Serving:} We use vLLM~\citep{kwon2023efficient} to generate outputs from our 8B models (\textsc{Aya} and \textsc{Qwen}), loading them with FP8 quantization and a maximum sequence length of 8,192 tokens. For the larger models (Command A and Gemini 2.0 Flash), we use their dedicated hosted APIs. For greedy decoding, we set top-$k$ to 1, while for multi-sample generation we obtain five completions at specified temperature and min-$p$ values.

\section{Choosing Judge and Reward Model}\label{app:reward_judge_choice}

\begin{table}[ht]
    \centering
    \begin{tabular}{lc}
        \toprule
        Model\textbf{} & \textbf{Avg} \\
        \midrule
        GPT-4o$^1$ (\texttt{gpt-4o-2024-08-06}) & 81.1\\
        GPT-4o$^2$ (\texttt{gpt-4o-2024-11-20}) & 85.8 \\
        \hdashline
        Aya Expanse 8B$^1$ & 65.2  \\
        Llama 3.1 70B$^1$ & 75.5 \\
        Gemma 2 9B$^1$ & 76.6 \\
        M-Prometheus-14B$^2$ & 79.5 \\
        Qwen2.5-14B-Instruct$^2$ & 80.8 \\
        Command A (111B) & 84.5 \\
        \bottomrule
    \end{tabular}
    \caption{Average accuracy on the M-RewardBench across 24 languages, comparing open models of various sizes (below) to GPT-4o variants. $^1$: Results are copied from ~\citep{gureja2024m}, $^2$: from ~\citep{pombal2025m}.}
    \label{tab:mrewardbench}
\end{table}

\begin{table}[ht]
    \centering
    \begin{tabular}{lc}
    \toprule
    \textbf{Language} & \textbf{Accuracy} \\
    \midrule
        arb\_Arab & 84.61\\
        ces\_Latn & 83.83\\
        deu\_Latn & 85.08 \\
        ell\_Grek & 84.04 \\
        fra\_Latn & 85.36 \\
        heb\_Hebr & 83.62\\
        hin\_Deva & 83.74\\
        ind\_Latn & 84.35 \\
        ita\_Latn & 85.90 \\
        jpn\_Jpan & 84.94 \\
        kor\_Hang & 83.43 \\
        nld\_Latn & 86.32 \\
        pes\_Arab & 81.98 \\
        pol\_Latn & 84.91 \\
        por\_Latn & 86.30 \\
        ron\_Latn & 83.22 \\
        rus\_Cyrl & 84.12\\
        spa\_Latn & 86.55\\
        tur\_Latn & 82.96\\
        ukr\_Cyrl & 83.83\\ 
        vie\_Latn & 84.49\\ 
        zho\_Hans & 84.59\\
        zho\_Hant & 84.25\\
        \midrule
        Avg & 84.45\\
    \bottomrule
    \end{tabular}
    \caption{Language breakdown of M-RewardBench scores for Command A.}
    \label{tab:commanda-mrewardbench}
\end{table}

\Cref{tab:mrewardbench} compares Multilingual RewardBench~\citep{gureja2024m} scores for multiple generative multilingual LLMs. We add benchmark scores for Command A by running the official code released with the benchmark\footnote{\url{https://github.com/Cohere-Labs-Community/m-rewardbench}, commit \texttt{5e5a0d3
}.}. Remaining scores are taken from prior reports ~\citep{gureja2024m,pombal2025m}.
\Cref{tab:commanda-mrewardbench} details the performance for Command A for each language. According to this benchmark and the scores reported in ~\citep{gureja2024m} and ~\citep{pombal2025m}, Command A is the best open judge, scoring closely to GPT-4o (and even outperforming an older variant).

There is further support in experiments by
~\citet{kreutzer2025d} where its agreement with pairwise human preferences from Chatbot Arena battles in multiple languages is close to GPT-4o's, with particular strengths in Chinese, Vietnamese, French, Turkish and Dutch.

On the English RewardBench benchmark~\citep{RewardBench},
classifier RMs are outperforming generative ones, so we pick the top-performing open model as our RM for BoN, \texttt{INF-ORM-Llama3.1-70B}~\cite{INF-ORM-Llama3.1-70B}, which is based on the multilingual Llama3.1-70B. The model underlying the RM, Llama3.1, supports English, German, French, Italian, Portuguese, Hindi, Spanish, and Thai---which we suspect yields the strong crosslingual generalization.

\section{Temperature Sensitivity}\label{app:temperatures}
\begin{figure*}[ht]
    \centering
    \includegraphics[width=0.8\linewidth]{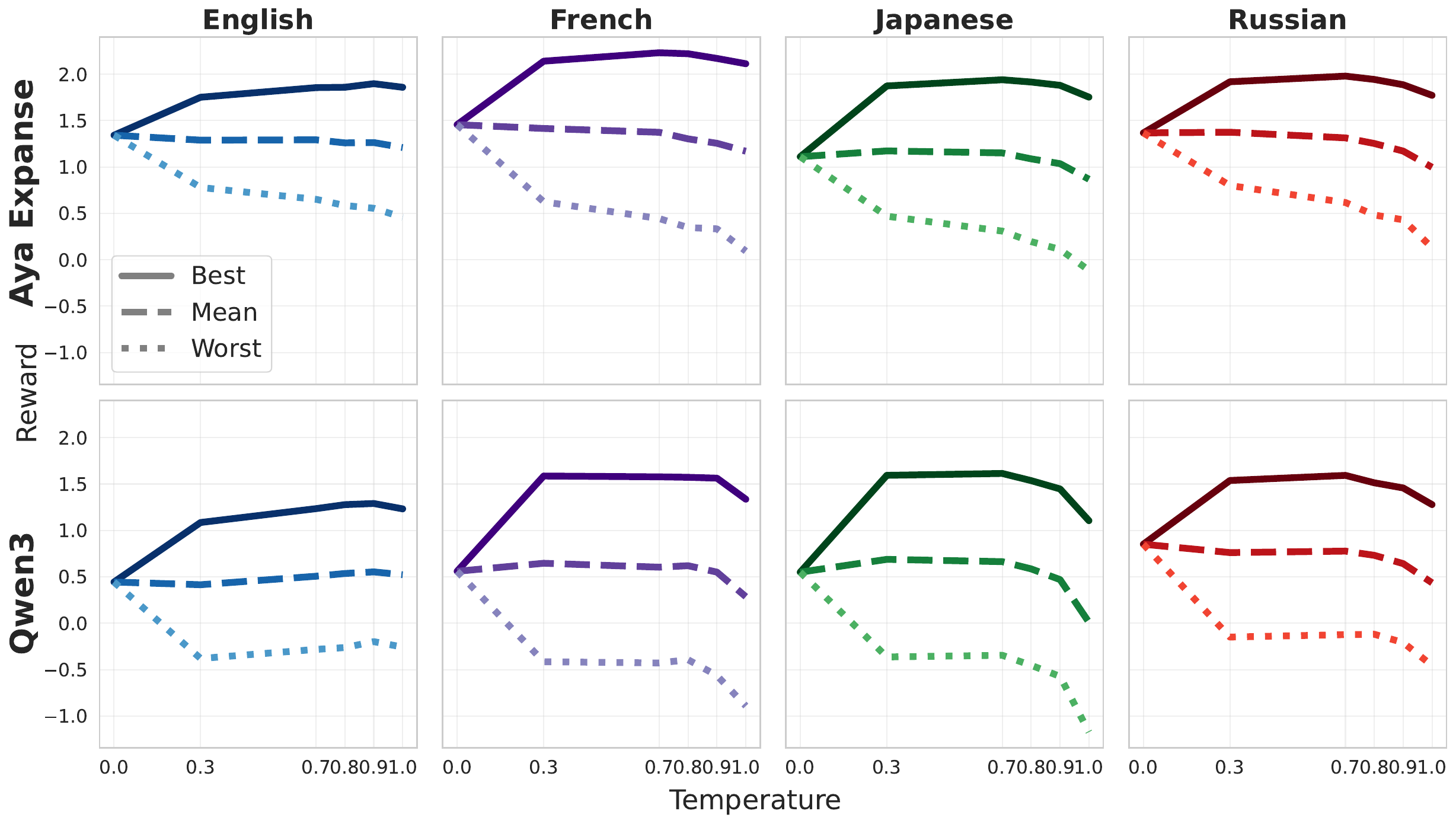}
    \caption{\textbf{Arena:} Evaluation score under different temperatures with $N=5$ samples.}
    \label{fig:temp_arena_more_langs}
\end{figure*}

\begin{figure*}[ht]
    \centering
    \includegraphics[width=0.8\linewidth]{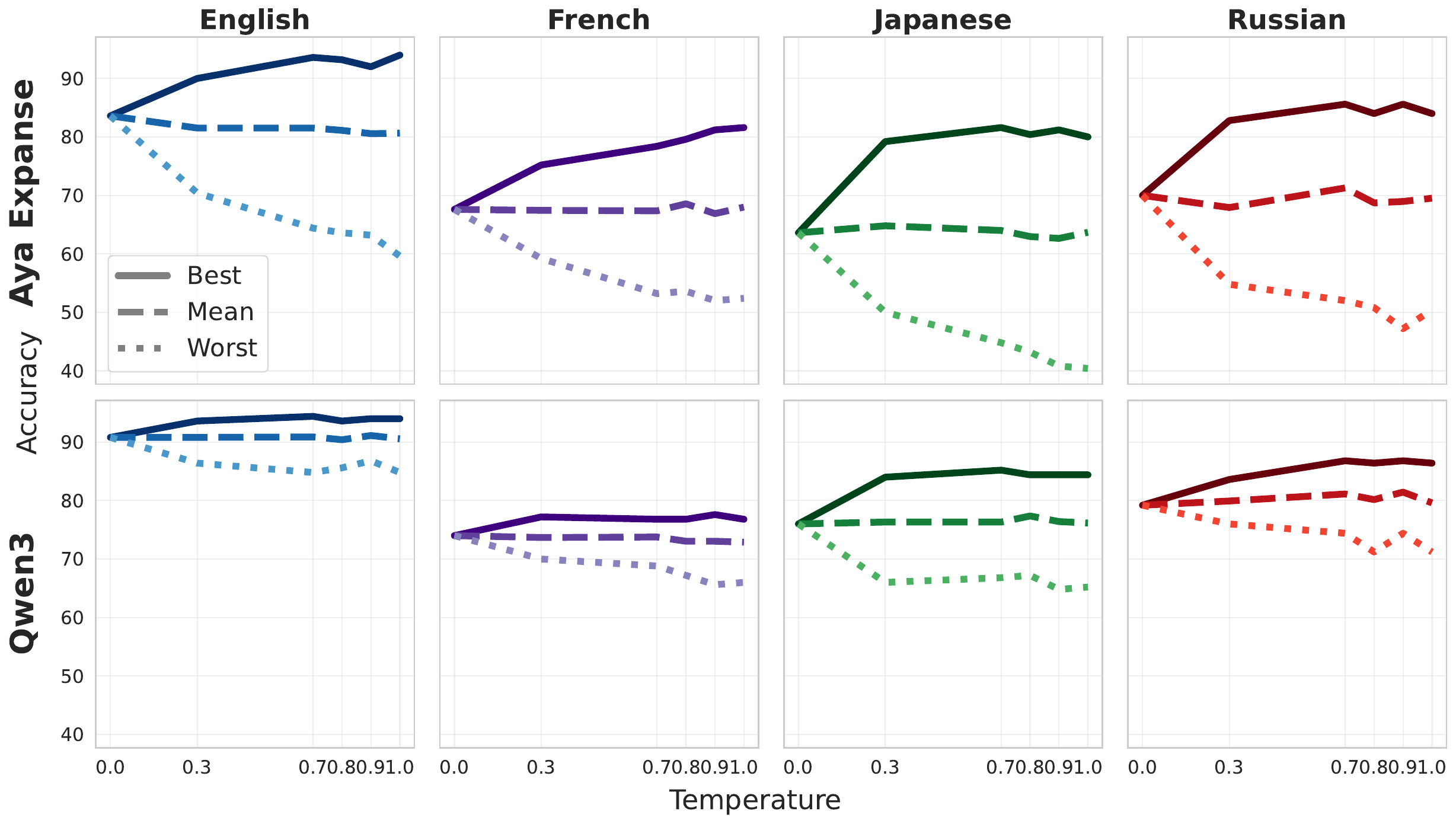}
    \caption{\textbf{MGSM:} Evaluation score under different temperatures with $N=5$ samples.}
    \label{fig:temp_mgsm}
\end{figure*}

\begin{figure*}[ht]
    \centering
    \includegraphics[width=0.8\linewidth]{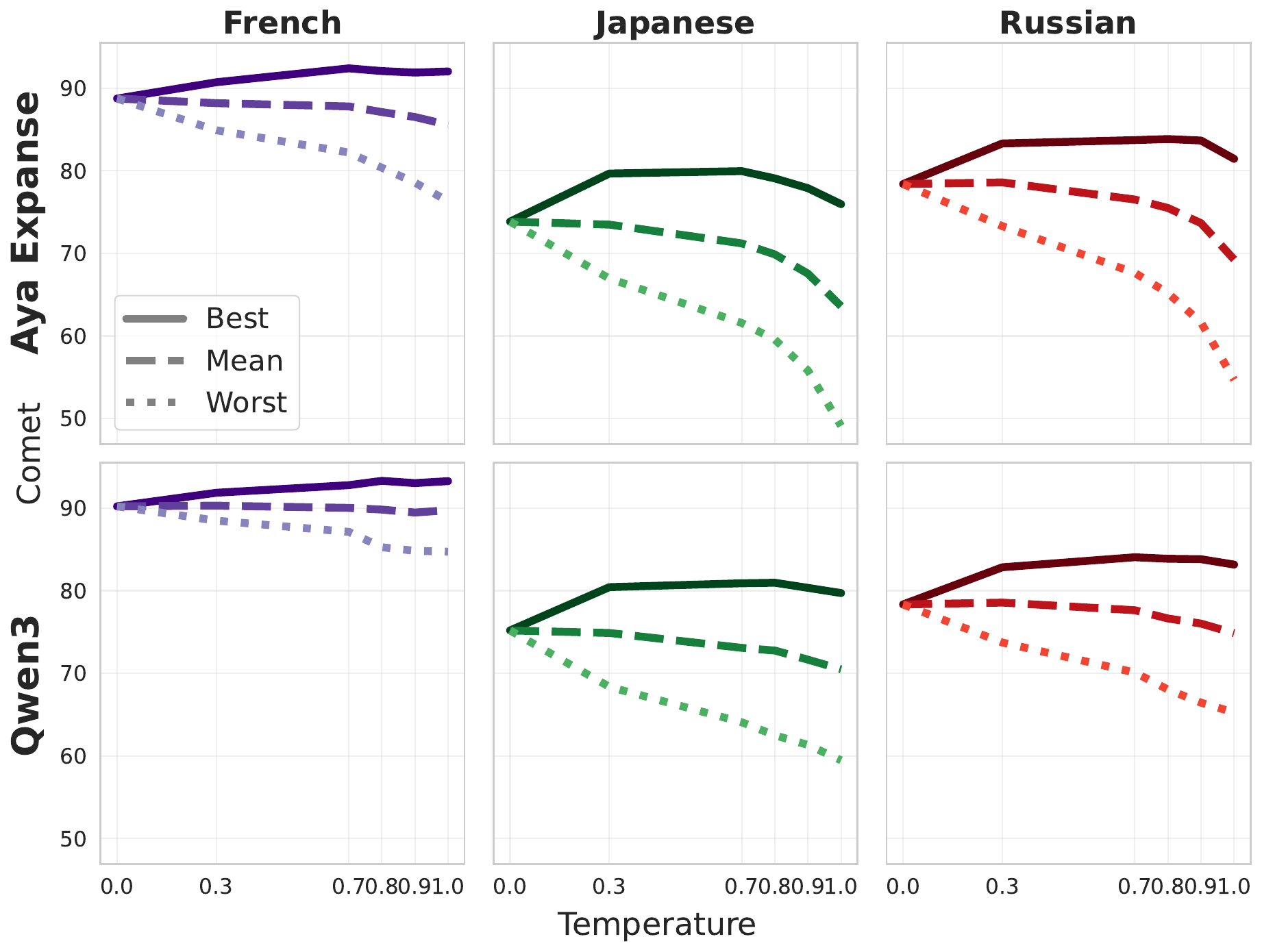}
    \caption{\textbf{WMT} Evaluation score under different temperatures with $N=5$ samples.}
    \label{fig:temp_wmt}
\end{figure*}

\Cref{fig:temp_arena_more_langs,fig:temp_mgsm,fig:temp_wmt} show the effect of temperature on the scores of best and worst generations and their mean on the dev splits under temperature sampling of varying temperatures for Arena, MGSM, and WMT, respectively. All consider $N=5$ samples.

\section{Selection Prompts}\label{app:prompts}
\Cref{lst:chops} reports the judge prompt for CHOPS,  \Cref{lst:xmbr} the one for X-MBR.

\begin{lstlisting}[
    float,
    floatplacement=ht,
    caption={Prompt Used for Checklisted One Pass Selection (CHOPS)},
    label={lst:chops}
]
"""
Please act as a fair judge. Based on the provided Instruction and Generated Texts in {language}, analyse the Generated Texts step by step and reply with the index of the best response.

(1) As a first step, write a tailored **evaluation checklist for the Instruction** that was given to an AI assistant. 
This evaluation checklist should be a list of questions that ask whether or not specific criteria relevant to the Instruction were met by an AI assistants response.
Each question in the checklist should measure one distinct aspect of quality.
Criteria covered by your checklist could be explicitly stated in the Instruction, or be generally sensible criteria for the problem domain.
Factuality is always of high importance, but where relevant, the checklist should also take language and culture-specific context into account. 
Questions should be concise and the checklist should not include unnecessary entries. Avoid vague wording and instead evaluate specific aspects of a response.

(2) As a second step, **compare the Generated Texts** with respect to the checklist. Write a brief explanation of the evaluation.

(3) In the third step, **output the index of the best Generated Text** according to the checklist evaluation.

## Output Format
Checklist: (each question should appear on a new line)
Q1: xxx
Q2: xxx

Explanation: xxx

Answer: [[ INDEX ]] (this should be an integer and nothing else; the index should be enclosed in double brackets as indicated)

## Evaluation Information

**Instruction**
{message}


**Generated Texts**
{generations}


Please analyse the Generated Texts according to a custom checklist and provide your selected text according to the guidelines. Remember to stick to the requested Output Format, providing the checklist questions and a short explanation and return the index of your selection inside double brackets [[]].
"""
\end{lstlisting}
\begin{lstlisting}[
    float,
    floatplacement=ht,
    caption={Prompt Used for Cross-Lingual MBR  (X-MBR)},
    label={lst:xmbr}
]

"""
Respond to the following instruction in {language_name}. Only return the answer in {language_name}.

{prompt}

Now Give your response  following the above instruction ONLY in {language_name}.        
"""
\end{lstlisting}

\section{Ablations}

\subsection{Choosing Single Temperatures}
One could tune the single temperature, but in practice, resources invested in such tuning might have limited return. For our example, we measure the maximum \textsc{hope} at $\tau=0.7$ of 48\% but it is close to the value at $\tau=0.8$ and $\tau=0.9$ of 47\% and 45\%, thus the effects of choosing one over the other might be negligible.

\subsection{Token-level hedging with min-$p$ sampling} \label{sec:min_p_ablations}
\begin{figure}[ht]
    \centering
    \begin{subfigure}[b]{0.48\textwidth}
        \centering
        \includegraphics[width=\textwidth]{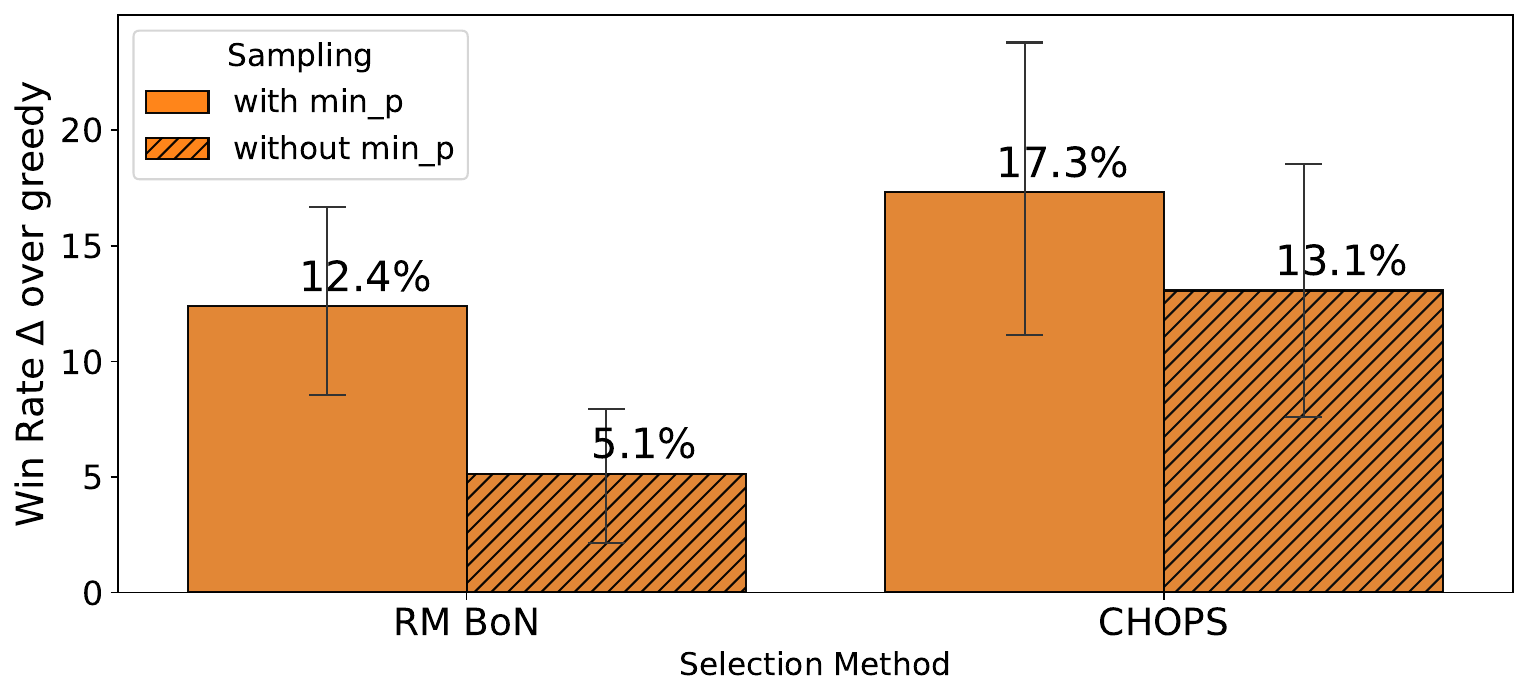}
        \caption{Multilingual win-rates.}
        \label{fig:min_p_winrates}
    \end{subfigure}
    \hfill
    \begin{subfigure}[b]{0.48\textwidth}
        \centering
        \includegraphics[width=\textwidth]{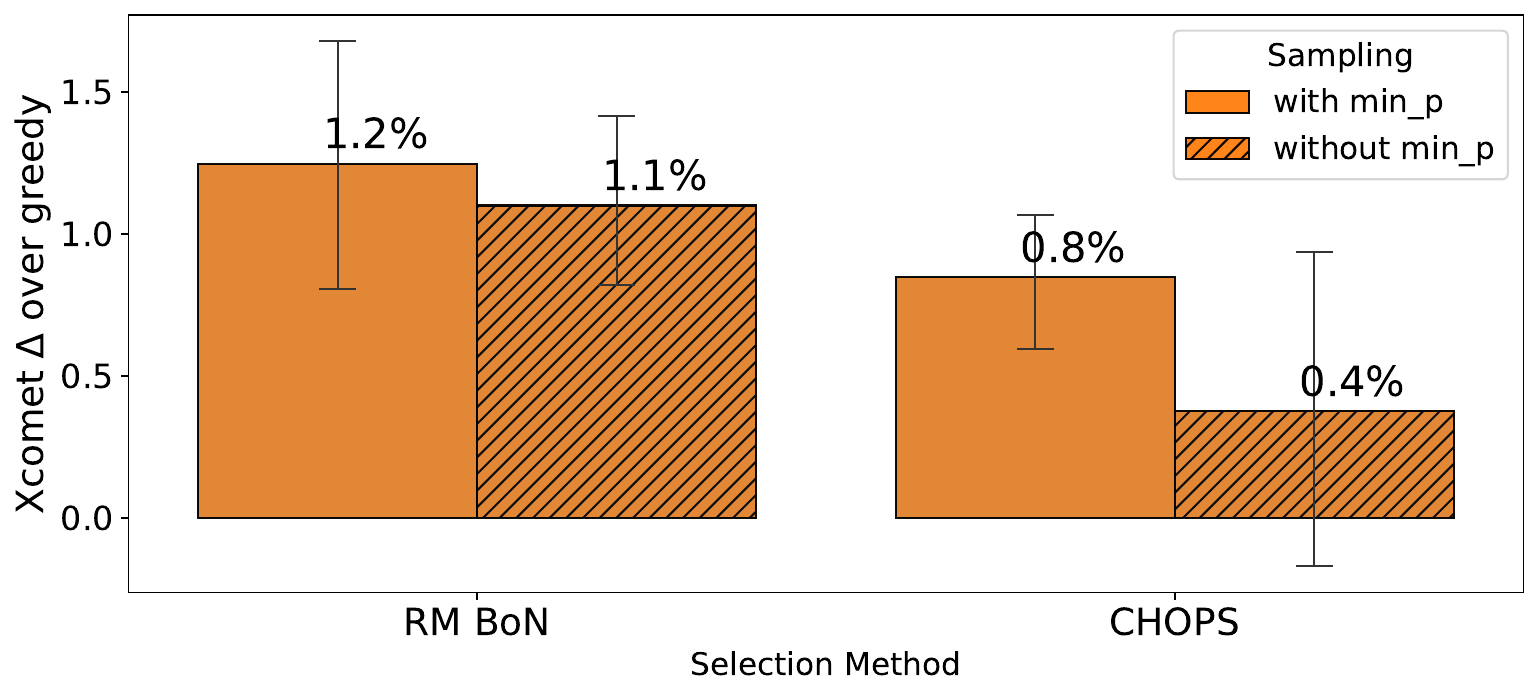}
        \caption{Machine translation}
        \label{fig:min_p_wmt}
    \end{subfigure}
    \caption{Effect of adding min-p sampling across different selection methods. Min-p provides consistent improvements across both methods and tasks. Results are on Arena (left) and WMT (right) dev splits, using \textsc{Aya}.}
    \label{fig:min_p_selection_effects}
\end{figure}

We consider min-$p$ sampling as an additional token-level hedging mechanism during generation. \Cref{fig:min_p_selection_effects} demonstrates that adding min-$p$ consistently improves performance across selection methods compared to only Hedged Sampling. For multilingual win-rates evaluation, min-$p$ provides substantial improvements for both RM BoN and CHOPS, while machine translation tasks show more modest but consistent benefits.

\subsection{One-Pass Selection} \label{sec:chops_ablation}
\begin{figure}[htbp]
  \centering
  \begin{subfigure}[b]{0.45\linewidth}
    \centering
    \includegraphics[width=\linewidth]{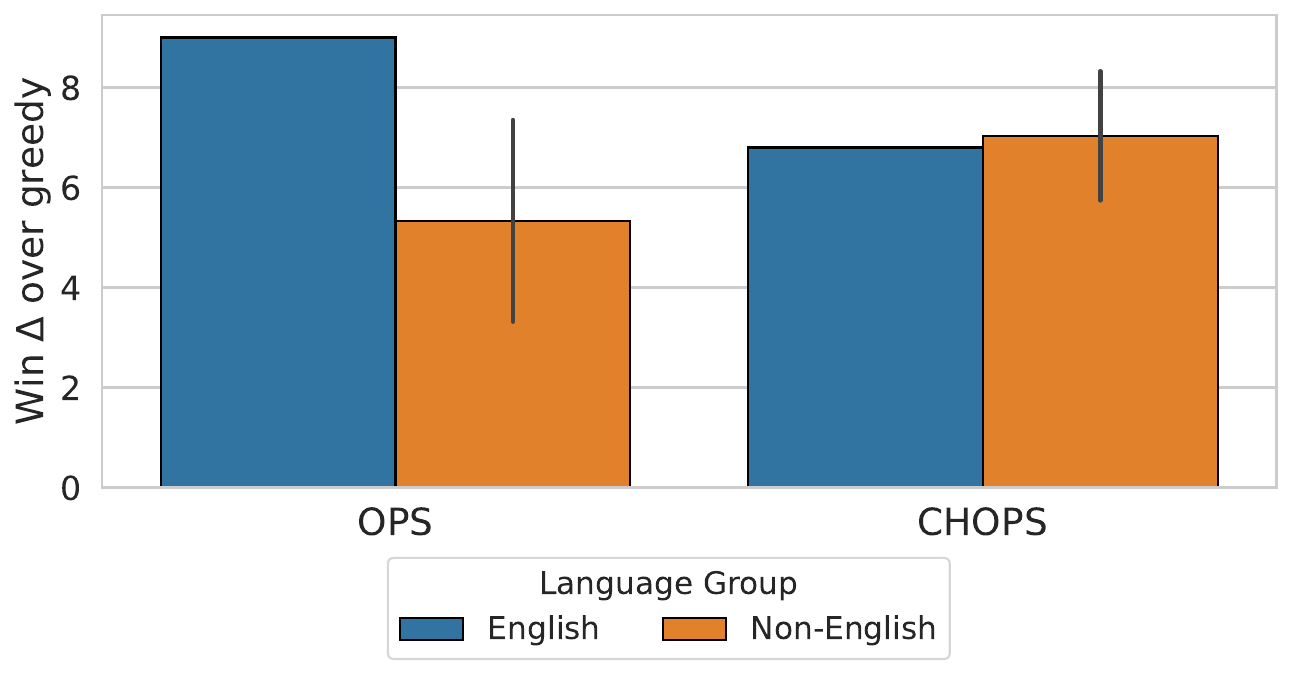}
    \caption{Win-rate delta on the Arena open-ended benchmark }
    \label{fig:chops_ablation_arena}
  \end{subfigure}
  \quad
  \begin{subfigure}[b]{0.45\linewidth}
    \centering
    \includegraphics[width=\linewidth]{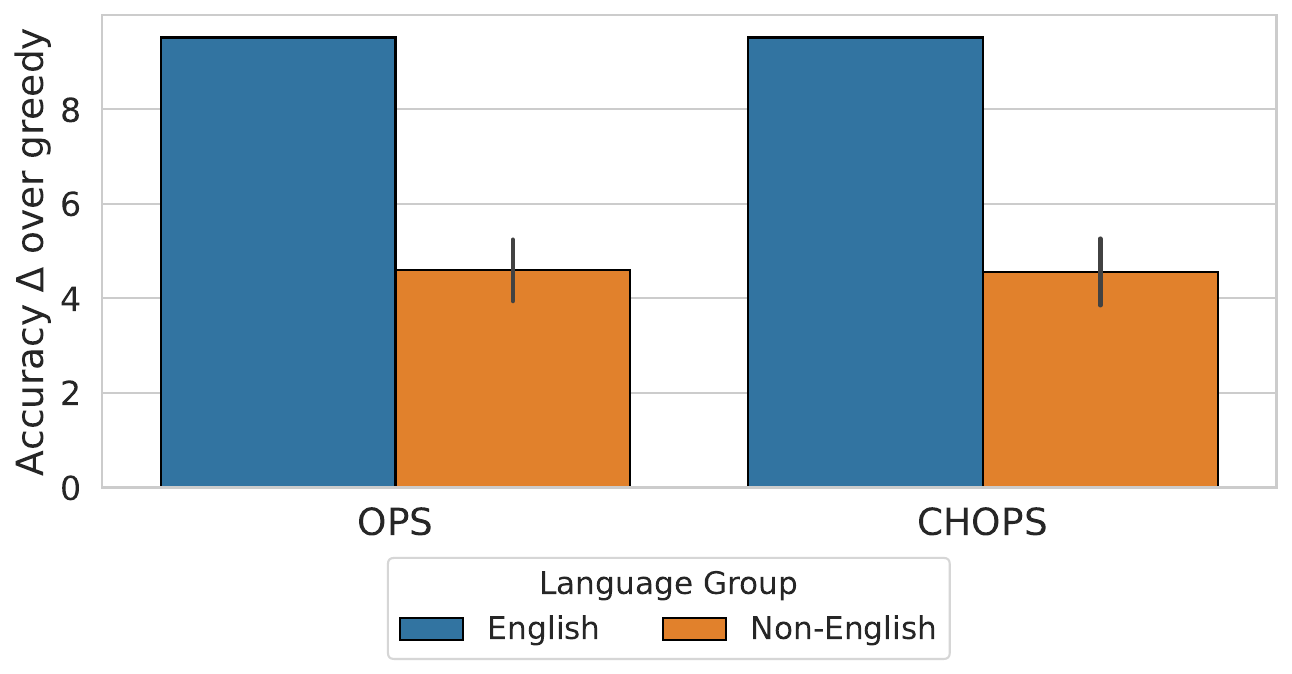}
    \caption{Accuracy on the GSM8K math benchmark}
    \label{fig:chops_ablation_mgsm}
  \end{subfigure}
  \caption{Comparison of CHOPS (with checklists) versus OPS (without checklists): (Left) The self-generated checklists improve multilingual performance on open-ended evaluation; (Right) Checklists have negligible effect on close-ended MGSM tasks. Results are averaged across models, using $N=5$ hedged samples and evaluated on dev splits.}
  \label{fig:chops_ablation}
\end{figure}

\Cref{fig:chops_ablation} compares our one‐pass, checklisted selection method (CHOPS) against a simpler one-pass selection (OPS) setup that chooses without any grounding checklist. We report the average win-rate delta over greedy decoding for English and non-English languages on the m-ArenaHard dev set in \Cref{fig:chops_ablation_arena}. OPS achieves a high win rate in English (9.0\%) but drops substantially in non-English (5.3\%), whereas CHOPS gives a more balanced outcome of 6.8\% and 7.1\% win-rates over greedy in English and non-English languages respectively. We hypothesize that this shift comes form the fact that generating a task-specific checklist before selection helps ground the judge’s decision process, reducing the effects of the variance in the sample and improving cross-lingual robustness. Notably, in \Cref{fig:chops_ablation_mgsm} the benefit of check-listed is less pronounced in verifiable tasks where comparison criteria are more explicitly defined.

\subsection{Sample Size Scaling}\label{sec:sample-scaling}
\begin{figure}
    \centering
    \includegraphics[width=0.6\linewidth]{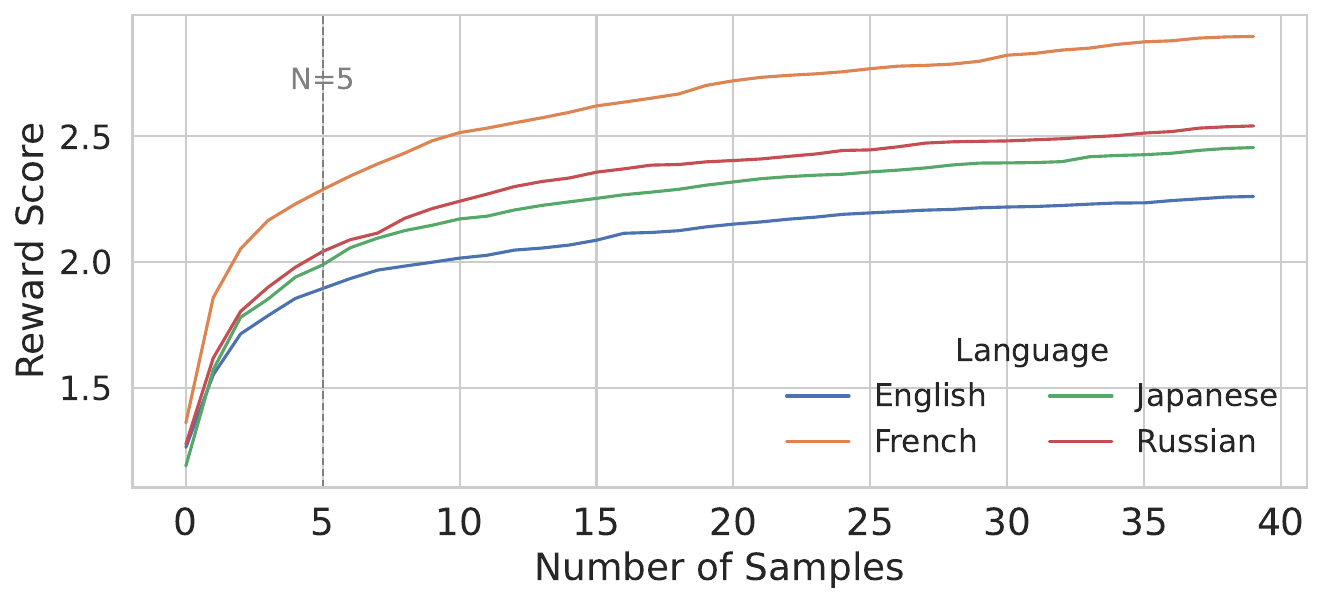}
\caption{BoN Reward score as we increase the sample size from from 1 sample to 40 samples}
\label{fig:scaling_rewards}
\end{figure}

\begin{figure*}
    \centering
    \includegraphics[width=1.0\linewidth]{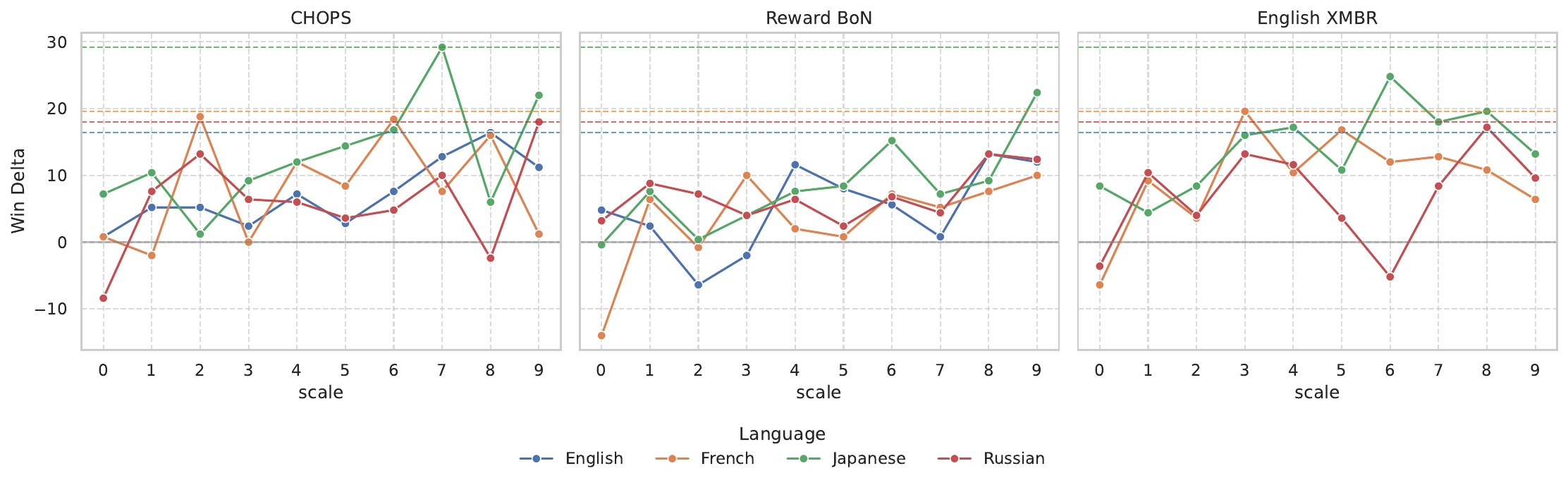}
\caption{\textbf{Scaling pool sample size} using $\tau=0.7$ hedged sampling for selected languages with different selection methods on mArenaHard dev set.}
\label{fig:scaling_winrate}
\end{figure*}

 We compare reward scores for mArenaHard when sampling ($\tau=0.7$, hedged) from \textsc{Aya} (on m-ArenaHard dev split) beyond the five times we have focused our experiments on. When reward and selection metric are perfectly aligned, such as in \cref{fig:scaling_rewards}, there is potential for improvement up till or even beyond $N=40$ for all languages. However, given that we are working with imperfect selection methods, that might not be perfectly aligned with the evaluation metric (e.g. MBR Judge vs accuracy), we do not expect these to transfer to realistic use cases. This was established theoretically by \citet{huang2025best} and \citet{stroebl2024inference}. The results in \Cref{fig:scaling_winrate} illustrates this issue: Win rate improvements over greedy are not developing smoothly across languages, sometimes even dropping below zero, so that it is not always the case that sampling more will result in a larger improvement in win rate. We can observe that RM BoN is most reliable in terms of its expected improvement with more samples, likely because it evaluates all samples in isolation and is thereby less affected by sample artifacts that the selection LLM could get misled by in pairwise (MBR) or direct comparisons (CHOPS).

\section{Evaluation Results}\label{app:results}
\Cref{tab:baselines_breakdown} shows the breakdown of baseline comparisons on the development set for hedged sampling ($\tau=0.7$, $N=5$).

\Cref{tab:selection_method_scores} contains the breakdown into individual languages and tasks for the test set evaluations of hedged sampling ($\tau=0.7$, $N=5$).

\begin{table*}[t]
\centering
\resizebox{0.8\textwidth}{!}{
\begin{tabular}{lllcccc}
\toprule
Model & Task & Language  & Likelihood & Sim. MBR & BoN & Judge MBR \\
\midrule
\multirow{20}{*}{\rotatebox{90}{\textsc{Aya}}} 
  & \multirow{7}{*}{Arena} 
    & Chinese & 4.00 & -6.40 & 5.20 & 4.00 \\
  &  & English & -7.60 & 10.40 & 12.00 & 11.20 \\
  &  & French & -15.20 & 0.40 & 14.80 & 1.60 \\
  &  & German & -12.80 & -1.20 & 17.20 & 6.40 \\
  &  & Japanese & 0.80 & -1.20 & 16.40 & 20.80 \\
  &  & Russian & -6.80 & 2.80 & 11.60 & 11.20 \\
  &  & Spanish & -6.00 & 3.60 & 4.00 & 4.80 \\
\cmidrule(lr){2-7}
  & \multirow{7}{*}{MGSM} 
    & Chinese & 1.76 & 2.96 & 12.16 & 9.36 \\
  &  & English & 13.92 & 16.72 & 13.92 & 17.92 \\
  &  & French & 2.16 & 2.56 & 4.56 & 4.96 \\
  &  & German & 0.80 & 2.00 & 6.40 & 4.80 \\
  &  & Japanese & -0.32 & 4.88 & 7.68 & 8.88 \\
  &  & Russian & -0.08 & 0.72 & 7.52 & 3.52 \\
  &  & Spanish & -0.08 & 1.52 & 6.72 & 4.72 \\
\cmidrule(lr){2-7}
  & \multirow{6}{*}{WMT} 
    & Chinese & 3.28 & 1.06 & 0.92 & -0.07 \\
  &  & French & 0.61 & -0.07 & 1.04 & -0.09 \\
  &  & German & 0.42 & 0.39 & 0.32 & 0.04 \\
  &  & Japanese & -0.19 & 0.10 & -0.18 & 0.11 \\
  &  & Russian & 0.43 & -0.07 & 1.08 & 0.15 \\
  &  & Spanish & -0.08 & -0.42 & 1.49 & 0.08 \\
\midrule
\multirow{20}{*}{\rotatebox{90}{\textsc{Qwen}}}
  & \multirow{7}{*}{Arena} 
    & Chinese & -1.60 & -2.40 & 3.20 & 7.20 \\
  &  & English & -7.20 & 0.80 & 0.80 & 3.20 \\
  &  & French & 1.60 & -1.20 & 1.60 & 3.20 \\
  &  & German & -7.60 & 7.20 & -4.00 & 5.20 \\
  &  & Japanese & -6.00 & -1.20 & 7.20 & 4.00 \\
  &  & Russian & -1.60 & -2.40 & -1.60 & 12.40 \\
  &  & Spanish & -4.00 & -1.20 & 8.80 & 16.80 \\
\cmidrule(lr){2-7}
  
  & \multirow{7}{*}{MGSM} 
    & Chinese & -1.44 & 0.96 & 3.36 & 2.16 \\
  &  & English & -0.88 & -0.48 & 2.32 & 0.32 \\
  &  & French & -1.20 & 0.40 & 2.00 & 1.60 \\
  &  & German & 0.72 & 1.12 & 2.72 & 1.52 \\
  &  & Japanese & 1.12 & 1.12 & 5.12 & 2.72 \\
  &  & Russian & 1.28 & 0.48 & 4.48 & 1.68 \\
  &  & Spanish & 0.00 & -0.40 & 0.40 & 0.80 \\
\cmidrule(lr){2-7}
  & \multirow{6}{*}{WMT} 
    & Chinese & 0.26 & -0.08 & 1.28 & 0.53 \\
  &  & French & 0.14 & -0.28 & 1.17 & 0.52 \\
  &  & German & 0.80 & 0.27 & 1.99 & 1.37 \\
  &  & Japanese & 0.51 & 0.17 & 1.941 & 1.61 \\
  &  & Russian & 0.68 & 0.26 & 2.27 & 1.11 \\
  &  & Spanish & 0.50 & 0.09 & 1.37 & 0.68 \\
\bottomrule
\end{tabular}
}

\caption{Breakdown of dev set results: Baseline Methods}
\label{tab:baselines_breakdown}
\end{table*}

\begin{table*}
\begin{tabular}{llcccccc}
\toprule
 &  &  & CHOPS & Judge MBR & Reward BoN & X-MBR & Greedy \\
Model & Task & Language &  &  &  &  &  \\
\midrule
\multirow{20}{*}{\rotatebox{90}{\textsc{Aya}}}  & \multirow[t]{7}{*}{Arena} & Chinese & 18.40 & 17.20 & 7.60 & 6.80 & -- \\
 &  & English & 13.60 & 6.40 & 5.20 & 13.20 & -- \\
 &  & French & 13.20 & 7.60 & 8.40 & 6.80 & -- \\
 &  & German & 11.60 & 5.20 & 0.80 & 16.40 & -- \\
 &  & Japanese & 9.20 & 4.00 & -0.40 & 10.40 & -- \\
 &  & Russian & 2.00 & -5.20 & 8.80 & 12.40 & -- \\
 &  & Spanish & 24.00 & 12.00 & 5.60 & 8.00 & -- \\
\cline{2-8}
 & \multirow[t]{7}{*}{MGSM} & Chinese & 8.56 & 7.36 & 13.96 & 12.16 & 68.64 \\
 &  & English & 7.76 & 7.36 & 13.16 & 9.36 & 77.84 \\
 &  & French & 9.04 & 7.84 & 13.84 & 10.24 & 62.96 \\
 &  & German & 5.12 & 5.92 & 10.32 & 6.72 & 73.68 \\
 &  & Japanese & 6.56 & 6.16 & 10.36 & 9.36 & 68.64 \\
 &  & Russian & 6.08 & 5.68 & 11.68 & 8.48 & 73.12 \\
 &  & Spanish & 8.96 & 8.56 & 12.96 & 9.76 & 70.24 \\

\cline{2-8}
 & \multirow[t]{6}{*}{WMT} & Chinese & 0.53 & 0.10 & 2.59 & -1.05 & 76.09 \\
 &  & French & 0.39 & 0.09 & 2.55 & -2.14 & 79.91 \\
 &  & German & 0.17 & 0.04 & 0.86 & -1.05 & 91.88 \\
 &  & Japanese & 1.12 & 0.65 & 3.03 & -3.62 & 78.99 \\
 &  & Russian & 0.60 & 0.31 & 2.86 & -2.73 & 82.51 \\
 &  & Spanish & 0.28 & 0.01 & 1.68 & -0.74 & 87.90 \\

\cline{1-8} \cline{2-8}
\multirow{20}{*}{\rotatebox{90}{\textsc{Qwen}}} & \multirow[t]{7}{*}{Arena} & Chinese & 0.80 & -4.00 & 8.80 & 13.60 & -- \\
 &  & English & 7.20 & 4.80 & 12.80 & 9.60 & -- \\
 &  & French & 8.00 & 1.20 & 8.40 & 18.40 & -- \\
 &  & German & 8.80 & -2.40 & -0.40 & 12.00 & -- \\
 &  & Japanese & 8.80 & 6.40 & 8.80 & 4.40 & -- \\
 &  & Russian & 8.00 & 2.40 & 3.60 & 12.00 & -- \\
 &  & Spanish & 8.00 & 4.80 & 16.00 & 4.80 & -- \\
\cline{2-8}
 & \multirow[t]{7}{*}{MGSM} & Chinese & 3.04 & 3.04 & 6.44 & 4.24 & 83.76 \\
 &  & English & -0.88 & -0.88 & 2.12 & -0.08 & 95.68 \\
 &  & French & 2.48 & 2.08 & 4.28 & 2.88 & 79.92 \\
 &  & German & 2.56 & 2.16 & 4.56 & 4.96 & 87.04 \\
 &  & Japanese & 3.28 & 3.68 & 6.48 & 4.88 & 80.72 \\
 &  & Russian & 1.36 & 2.56 & 3.96 & 2.96 & 89.04 \\
 &  & Spanish & 2.48 & 2.48 & 5.28 & 4.48 & 86.72 \\
\cline{2-8}
 & \multirow[t]{6}{*}{WMT} & Chinese & 0.51 & -0.29 & 2.18 & -0.23 & 78.42 \\
 &  & French & 1.75 & 0.06 & 2.81 & 0.09 & 77.71 \\
 &  & German & 0.87 & -0.43 & 1.56 & -0.23 & 89.86 \\
 &  & Japanese & 1.69 & -0.68 & 3.75 & -0.60 & 74.36 \\
 &  & Russian & 2.21 & -1.68 & 3.86 & -1.26 & 77.64 \\
 &  & Spanish & 1.21 & -0.36 & 2.24 &0.26 & 85.96 \\
\cline{1-8} \cline{2-8}

\bottomrule
\end{tabular}

\caption{Breakdown of devtest set results: Judge based Methods}
\label{tab:selection_method_scores}
\end{table*}

\end{document}